%% file: main.tex
\documentclass[10pt,twocolumn,letterpaper]{article}
\usepackage{graphicx}
\graphicspath{ {./figs/} }

\usepackage[pagenumbers]{cvpr} 
\usepackage{multirow}
\input{preamble}

\definecolor{cvprblue}{rgb}{0.21,0.49,0.74}
\usepackage[pagebackref,breaklinks,colorlinks,citecolor=cvprblue]{hyperref}

\usepackage[normalem]{ulem} 
\usepackage{xcolor} 

\usepackage{titling} 

\usepackage{float}
\usepackage{floatpag}
\usepackage{afterpage}
\usepackage{subcaption}
\usepackage[margin=1in]{geometry}
\usepackage{amsmath}
\usepackage{bm}
\usepackage{stfloats}
\usepackage{pgfplots}
\pgfplotsset{compat=1.17}
\usepackage{tikz}
\usepackage{overpic}
\usepackage{cuted}
\usepackage{capt-of}

\usepackage[linesnumbered,ruled]{algorithm2e}

\newcommand{\ee}[1]{\textsc{EmbEdit}}

\title{Implicit Priors Editing in Stable Diffusion via Targeted Token Adjustment}

\author{
Feng He\\
University of Sheffield, UK\\
{\tt\small fhe12@sheffield.ac.uk}
\and
Chao Zhang\\
Toshiba Europe Ltd, UK\\
{\tt\small chao.zhang@toshiba.eu}
\and
Zhixue Zhao\\
University of Sheffield, UK\\
{\tt\small zhixue.zhao@sheffield.ac.uk}
}

\begin{document}
\maketitle

\begin{strip}
  \centering
   \includegraphics[width=0.9411\textwidth]{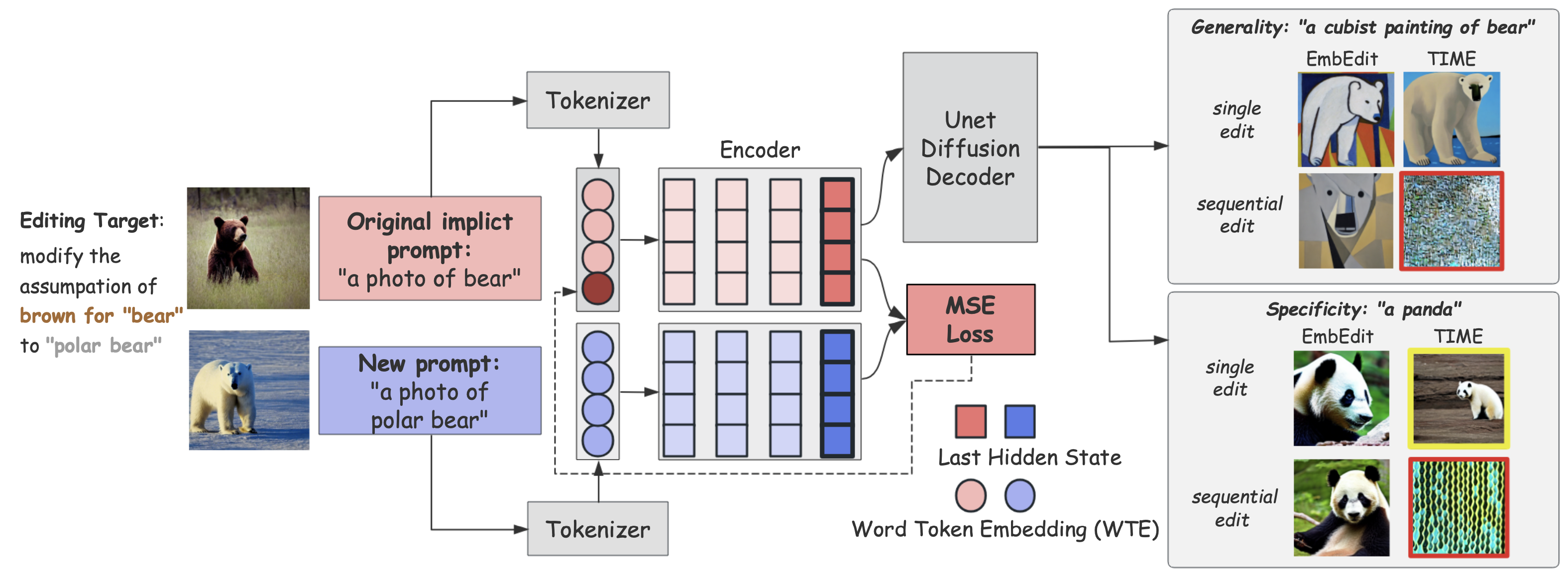}
   \captionof{figure}{Illustration of the proposed \ee{} modifying the word-token embedding (WTE) of the target word ``bear'' to shift its interpretation from ``brown bear" to ``polar bear". \ee{} optimizes the WTE of ``bear'' by minimizing the distance between the last hidden states of the text encoder in Stable Diffusion for both the original implicit prompt and the explicit prompt. Since \ee{} keeps the model weights completely unchanged, it supports sequential editing without performance degradation or model collapse, as shown in the red-bordered example images. Furthermore, \ee{} does not modify unrelated objects' WTE, preventing undesirable effects on unrelated objects, as demonstrated by TIME in the yellow-bordered box (where a panda head appears with a polar bear body).}
   \label{fig:embedit_illustrate}
\end{strip}

\input{sec/0_abstract}

\input{sec/1_intro}
\input{sec/2_related_probe}

\input{sec/3_method}
\input{sec/4_eval}
\input{sec/5_gender}
\input{sec/6_conclusion}
{
    \small
    \bibliographystyle{ieeenat_fullname}
    \bibliography{main}
}

\clearpage
\setcounter{section}{0} 
\renewcommand{\thesection}{\Alph{section}}
\renewcommand{\thefigure}{\thesection.\arabic{figure}}
\renewcommand{\thetable}{\thesection.\arabic{table}}
\setcounter{figure}{0}
\setcounter{table}{0}

\input{sec/X_suppl}

\end{document}

%% file: preamble.tex
\usepackage[dvipsnames]{xcolor}


%% file: sec/0_abstract.tex
\begin{abstract}






Implicit assumptions and priors are often necessary in text-to-image generation tasks, especially when textual prompts lack sufficient context. However, these assumptions can sometimes reflect outdated concepts, inaccuracies, or societal bias embedded in the training data.
We present Embedding-only Editing (\ee{}), a method designed to efficiently adjust implict assumptions and priors in the model without affecting its interpretation of unrelated objects or overall performance. Given a ``source" prompt (e.g., ``rose") that elicits an implicit assumption (e.g., rose is red) and a "destination" prompt that specifies the desired attribute (e.g., ``blue rose"), 
\ee{} fine-tunes only the word token embedding (WTE) of the target object (``rose") to optimize the last hidden state of text encoder in Stable Diffusion, a SOTA text-to-image model.
This targeted adjustment prevents unintended effects on other objects in the model's knowledge base, as the WTEs for unrelated objects and the model weights remain unchanged. Consequently, when a prompt does not contain the edited object, all representations, and the model outputs are identical to those of the original, unedited model.
Our method is highly efficient, modifying only 768 parameters for Stable Diffusion 1.4 and 2048 for XL in a single edit, matching the WTE dimension of each respective model. This minimal scope, combined with rapid execution, makes \ee{} highly practical for real-world applications. Additionally, changes are easily reversible by restoring the original WTE layers.
Our experimental results demonstrate that \ee{} consistently outperforms previous methods across various models, tasks, and editing scenarios (both single and sequential multiple edits), achieving at least a 6.01\% improvement (from 87.17\% to 93.18\%).


\end{abstract}

%% file: sec/1_intro.tex
\section{Introduction}\label{sec:intro}



\begin{figure*}[b!]
    \vspace{0.13cm}
    {\textbf{Edit ``cake" to ``red velvet cake"}} 
    \vspace{0.2cm}
    \centering
    \vspace{-0.15cm}
    \resizebox{\textwidth}{!}{%
        \begin{minipage}[b]{0.33\textwidth}  
        \centering
        \includegraphics[width=0.3\linewidth]{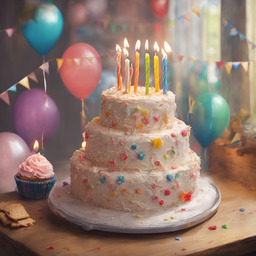}
            \hspace{-5pt} 
            \begin{overpic}[width=0.3\linewidth]{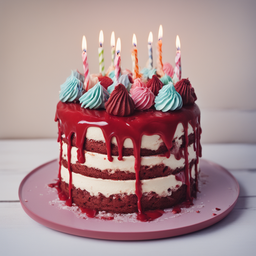}
                \put(-35, 20){\includegraphics[width=0.2\linewidth]{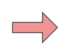}}
            \end{overpic}
            \hspace{-3pt}
            \includegraphics[width=0.3\linewidth]{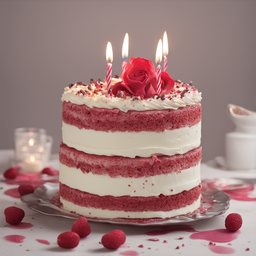}
            \vspace{-8pt}
            \captionof*{figure}{\footnotesize ``a birthday cake''}
        \end{minipage}
        
        \begin{minipage}[b]{0.33\textwidth}
            \centering
            \includegraphics[width=0.3\linewidth]{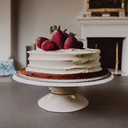}
            \hspace{-5pt}
            \begin{overpic}[width=0.3\linewidth]{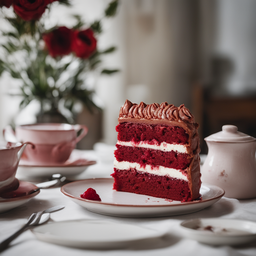}
                \put(-35, 20){\includegraphics[width=0.2\linewidth]{figs/right_arrow_1.png}}
            \end{overpic}
            \hspace{-3pt}
            \includegraphics[width=0.3\linewidth]{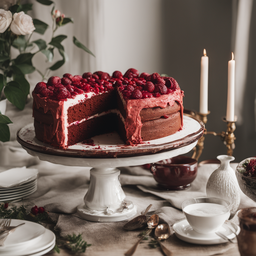}
            \vspace{-8pt}
            \captionof*{figure}{\footnotesize ``a photo of cake on the dining table''}
        \end{minipage}

        \begin{minipage}[b]{0.33\textwidth}
            \centering
            \includegraphics[width=0.3\linewidth]{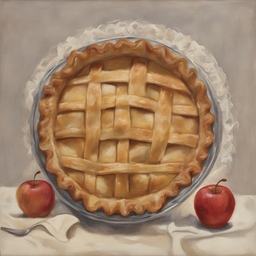}
            \hspace{-5pt}
            \begin{overpic}[width=0.3\linewidth]{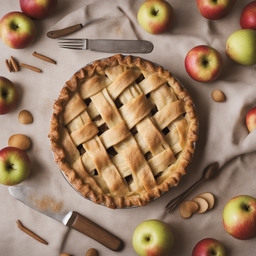}
                \put(-35, 20){\includegraphics[width=0.2\linewidth]{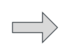}}
            \end{overpic}
            \hspace{-3pt}
            \includegraphics[width=0.3\linewidth]{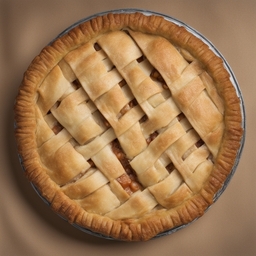}
            \vspace{-8pt}
            \captionof*{figure}{\footnotesize ``an apple pie''}
        \end{minipage}
    }
    
    \vspace{0.1cm}
    {\textbf{Edit ``chair" to ``massage chair"}}
    \vspace{-.3cm}
    \resizebox{\textwidth}{!}{%
        \begin{minipage}[b]{0.33\textwidth}  
            \centering
            \includegraphics[width=0.3\linewidth]{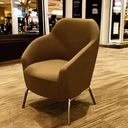}
            \hspace{-5pt} 
            \begin{overpic}[width=0.3\linewidth]{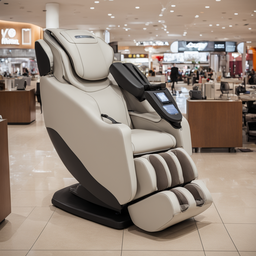}
                \put(-35, 20){\includegraphics[width=0.2\linewidth]{figs/right_arrow_1.png}}
            \end{overpic}
            \hspace{-3pt}
            \includegraphics[width=0.3\linewidth]{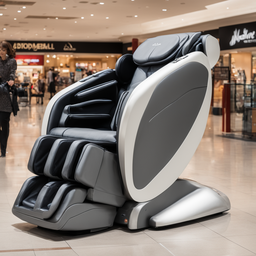}
            \vspace{-8pt}
            \captionof*{figure}{\footnotesize ``a chair in a mall''}
        \end{minipage}
        
        \begin{minipage}[b]{0.33\textwidth}
            \centering
            \includegraphics[width=0.3\linewidth]{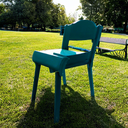}
            \hspace{-5pt}
            \begin{overpic}[width=0.3\linewidth]{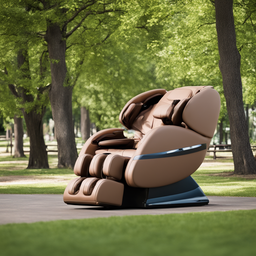}
                \put(-35, 20){\includegraphics[width=0.2\linewidth]{figs/right_arrow_1.png}}
            \end{overpic}
            \hspace{-3pt}
            \includegraphics[width=0.3\linewidth]{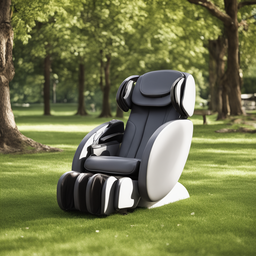}
            \vspace{-8pt}
            \captionof*{figure}{\footnotesize ``a chair in the park''}
        \end{minipage}
        
        \begin{minipage}[b]{0.33\textwidth}
            \centering
            \includegraphics[width=0.3\linewidth]{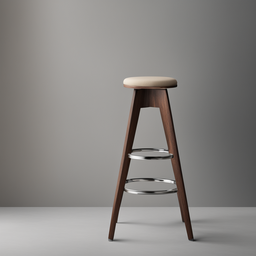}
            \hspace{-5pt}
            \begin{overpic}[width=0.3\linewidth]{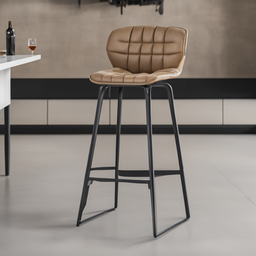}
                \put(-35, 20){\includegraphics[width=0.2\linewidth]{figs/right_arrow.png}}
            \end{overpic}
            \hspace{-3pt}
            \includegraphics[width=0.3\linewidth]{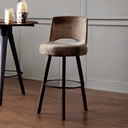}
            \vspace{-8pt}
            \captionof*{figure}{\footnotesize ``a bar stool''}
        \end{minipage}
    }
    \captionof{figure}{\small \ee{} edits implicit assumptions in text-to-image models. Two examples are shown here. In each row, columns 1 and 2 display positive prompts related to the target word, while column 3 shows the generated image for an unrelated object prompt.}
    \label{fig:overallpic}
\end{figure*}
Text-to-image models are powerful tools that generate visual content from textual descriptions, translating written prompts into detailed images. One of the most advanced among these is the Stable Diffusion model, renowned for its flexibility and accuracy in generating high-quality images~\cite{rombach2022high}.

Stable diffusion~\cite{rombach2022high} generates images based on a given text prompt. When the textual descriptions in prompts are ambiguous or lack essential details, the model fills in the gap with default, implicit priors. For example, the description ``an apple" may implicitly assume the color ``red''.
These implicit assumptions or priors help the model resolve ambiguities in under-specified prompts by drawing on common associations learned during training. However, such assumptions can introduce issues in certain contexts, as they may reflect social biases~\cite{Wan2023KellyIA,Haim2024WhatsIA,Shin2024AskLD,Wan2024WhiteML}, or outdated information~\cite{gandikota2024unified,arad-etal-2024-refact}. 
To address this issue, some approaches have focused on modifying model parameters to alter specific implicit assumptions, such as by adjusting the cross-attention layer
~\citep{orgad2023editing} or the MLP layers~\citep{arad-etal-2024-refact}. While these methods can effectively change certain assumptions, they typically require updating a subset of the model parameters within specific  components. Such parameter modifications, however, carry the risk of affecting other knowledge and associations that should remain intact, especially when multiple edits are applied.



Inspired by prior work on word embedding bias analysis~\citep{bolukbasi2016man,swinger2019biases,zhao-etal-2019-gender}, we propose \ee{}, which only modifies the Word Token Embeddings (WTE) of the target object to adjust the encoded priors, as illustrated in Figure~\ref{fig:embedit_illustrate}. Specifically, \ee{} modifies the WTE (i.e., the representations of the corresponding token ID) of the object token only, e.g., WTE of ``rose'', leaving representations of (i) non-target words, (ii) model parameters, and (iii) the diffusion modules unaffected. 
Importantly, our approach is \emph{side-effect free}: when the prompt does not contain the target object token, the inference operation for this prompt remains identical before and after \ee{}.
Our approach is also extremely \emph{parameter-efficient}: it does not fine-tune any modules of the model but only updates the embedding vector of the target WTE, which is a 768-dim vector and accounts for only 0.002\% of the Stable Diffusion v1.4. This leads to much less parameters comparing to previous methods, such as TIME~\citep{orgad2023editing}. TIME updates the cross-attention component, which accounts for 2.2\% of the model size. Moreover, \ee{} can be easily scaled to thousands of edits without performance degradation. This is attributed to the fact that the diffusion model remains intact.

We find that \ee{} achieves precise model editing with effective generalization 
in both single editing (only one target is edited) in Figure~\ref{fig:overallpic} and sequential editing modes (multiple targets are edited) in Figure~\ref{fig:compare_single_seq}. We illustrate the pipeline of \ee{} in Figure~\ref{fig:embedit_illustrate} for a prompt ``a photo of bear". Our \ee{}-edited Stable Diffusion v1.4 model generates a polar bear (Edit Efficacy) and successfully generalizes this concept to ``a cubist painting of a bear" (Edit Generality), while preserving the original representation of the unrelated term ``panda" (Edit Specificity). For comparison, we also present the outputs from TIME-edited Stable Diffusion v1.4, which produces distorted images of ``panda", showcasing undesirable side effects, i.e. poor specificity.
Our experiments on object editing and gender bias mitigation datasets show that \ee{} consistently outperform previous methods in achieving editing targets across different backbone models, edit counts, and tasks. We summarize our contribution as follows.
\begin{itemize}
    \item We probe the 
    color-related signals embedded in the WTE representation, and identify that WTE potentially contributes to implicit assumptions in the model (Sec \ref{sec:Emb_probing}). 
    \item We propose \ee{}, a novel model editing method for text-to-image models that updates only the WTE vector of the target object. 
    Compared to previous methods, \ee{} delivers superior results and is preferred for its ability to isolate edits to target words without affecting non-target words,  while maintaining parameter efficiency (Sec~\ref{sec:method}).
    \item We broaden the evaluation experiments by increasing both the number of concurrent edits and the model sizes, moving beyong the single-edit limitation and the focus on Stable Diffusion v1.4 in prior studies. \ee{} consistently outperforms state-of-the-art methods across various edit tasks, evaluation metrics, and model sizes (Sec~\ref{sec:experiment} and \ref{sec:gender}).

\end{itemize}


%% file: sec/2_related_probe.tex

\section{Related Works}

Stable Diffusion model is a popular state-of-the-art (stoa) Diffusion based Text-to-image (T2I) Models~\citep{rombach2022high}. It consists of two components: CLIP text encoder, converting the input text into latent text representation vectors~\citep{radford2021learning}, and diffusion model, taking the text representations and generating images by progressively reversing a noise process~\citep{sohl2015deep}. The CLIP text encoder is built upon Transformer~\cite{Vaswani2017AttentionIA} architecture, which processes the input token sequence through multi-head attention and feed-forward networks.





Similar to large language models (LLMs), T2I models encode knowledge and perceptions about the world, which can be inaccurate, outdated, or biased. Model editing, initially explored in LLMs, has shown a promising approach to control model behaviors post-training without extensive fine-tuning and data
curation~\cite{mitchell2022fast,hartvigsen2023aging,tan23malmen,meng2022locating, meng2022mass}.
Recent work has introduced several methods for editing text-to-image diffusion models, including erasing concepts~\citep{Lu_2024_CVPR,gandikota2024unified,basu2024localizing} and artistic styles~\citep{gandikota2024unified}, modifying implicit assumpations~\citep{orgad2023editing} and editing factual knowledge~\citep{arad-etal-2024-refact}. 

A key challenge in T2I editing is localizing knowledge within the model, which is crucial for subsequent modifications.
\cite{basu2024localizing} investigates the storage of distinct visual attributes in T2I diffusion models using Causal Mediation Analysis~\citep{10.1145/3501714.3501736,meng2022locating}. 
Their findings reveal that, unlike in LLMs where knowledge is typically concentrated in mid-MLP layers~\citep{meng2022locating, meng2022mass}, T2I models distribute attribute information across multiple components of the conditional UNet. This distributed nature prevents direct adaptation of LLM editing methods to T2I models and complicates the process of knowledge localization and modification.

Further, a fundamental challenge in T2I editing lies in achieving targeted modifications while keeping unrelated objects and concepts unchanged~\citep{orgad2023editing}. Existing methods modify model parameters, such as the $W_K$ and $W_V$ matrices for text input in cross-attention~\citep{orgad2023editing, Lu_2024_CVPR} or the $W_\text{project}$ matrix in the text encoder's MLP~\citep{arad-etal-2024-refact}. However, since representations of non-target objects must pass through and interact with these modified components, their inference process is altered in the edited model, leading to unintended changes in the output. This dilemma presents a critical trade-off between edit efficacy (achieving the desired modification), generality (applying the edit across diverse contexts), and specificity (a.k.a. locality - ensuring changes affect only the targeted concepts)~\citep{meng2022locating}.

\section{Probing Word Token Embedding}\label{sec:Emb_probing}
Previous work has highlighted the presence of biases in Word Token Embedding (WTE) in language models, particularly concerning sensitive attributes like race and gender~\citep{swinger2019biases,10.5555/3157382.3157584,zhao-etal-2019-gender,dong2023probing,gallegosetal2024bias,may2019measuring,abid2021persistent}. These biases can cascade through model architectures and manifest in downstream applications, potentially resulting in biased algorithmic outcomes~\citep{papakyriakopoulos2020bias,qian2022perturbation,garimella2022demographic,Venkit2023NationalityBI,Zayed2022DeepLO,Liu2021MitigatingPB,Gira2022DebiasingPL,Joniak2022GenderBA}.
Inspired by these findings, we hypothesize that the implicit biases observed in text-to-image models originate from text embeddings. Specifically, the biases come from the WTE in the CLIP text encoder. Therefore, in this section, we employ a probing task to validate this hypothesis. Given an object commonly in color red (e.g., ``apple") or yellow (``lemon"), we take its WTE of CLIP text encoder as the feature, and we set up a simple task to predict its color. Intuitively, high prediction accuracy would indicate that the WTE representations in CLIP's text encoder inherently encode color information, suggesting that object color presumptions are embedded directly within the WTE layer. Our hypothesis contrasts with prior work that locates such presumption in the cross-attention in the UNet decoder~\citep{orgad2023editing} or in the MLP layers in the text encoder~\citep{arad-etal-2024-refact}.

\paragraph{Probing Task}
A probing task is a diagnostic method often employed in computational linguistics to analyze the internal representations learned by a language model~\citep{ettinger-etal-2016-probing,eger-etal-2020-probe,sahin-etal-2020-linspector}. It involves training a simple classifier, a ``probe", on the model's embeddings or hidden layers to predict specific linguistic properties, such as numeracy~\citep{wallace-etal-2019-nlp}, hypernym~\citep{ravichander-etal-2020-systematicity}, sentence length~\citep{Conneau2018WhatYC}, or syntax~\citep{hewitt-manning-2019-structural}.
In our probing task, we use the WTE of an object as the feature to predict its color. 

To construct our probe task dataset, we query ChatGPT to generate a list of 200 common objects in red or yellow color: 100 red objects and 100 yellow objects. The color serves as the binary label (``red'' or ``yellow'') for each object. We then extract Word Token Embeddings (WTE) for these objects using the CLIP text encoder to obtain their feature representations. The dataset is split into an 80:20 ratio for training and testing sets.

\paragraph{Probing Classifier}

We train a logistic regression classifier to predict object colors based on text encoder from the CLIP model.
The probe achieves an accuracy of 90\%(±1.25\%) over five random seeds on the test set. Our further analysis of incorrect predictions reveals that errors arise from objects with ambiguous or variable colors in real-world contexts. For example, 
objects like ``clownfish" and ``sunsets" feature both red and yellow, making their color classification inconclusive. Overall, \textbf{the high classification accuracy of the probing task suggests that the WTEs of the CLIP text encoder effectively encode color signals.} Further details on the dataset, feature extraction, and accuracy calculation are provided in the supplementary materials.

%% file: sec/3_method.tex
\section{Our Method}\label{sec:method}

We extend the findings from Section~\ref{sec:Emb_probing} to additional visual attributes, such as shapes, sizes, and specific categories (``collie'' for dog). Our idea is to simply leverage the fact that WTEs encode implicit priors about different objects and concepts. The goal of \ee{} is to modify the target WTEs to adjust these encoded priors.

As shown in Figure~\ref{fig:embedit_illustrate} and Algorithm~\ref{alg:pseduo_code}, \ee{} locates and modifies the WTE of the object token, $\mathrm{wte}_\text{``bear"}\in \mathbb{R}^{768}$.
The optimization process minimizes the distance between textual representations of the original object token, $\mathbf{h}_\text{``bear"}$, and new object tokens with the target attribute, $\mathbf{h}_\text{``polar bear"}$. The representation $\mathbf{h}$ is the last hidden states of CLIP text encoder. By minimizing the MSE loss (Equ~\ref{eq:mse_loss}) between the hidden states, we aim to finetune $\mathrm{wte}_{\text{orig}}$ to reduce the semantic discrepancy between the original and new concepts. Here, we adopt the last hidden state $\mathbf{h}_{\text{orig}}$ and $\mathbf{h}_{\text{new}}$ as semantic rich representations of text prompts. To proceed, the model specifically updates and optimizes the WTE vector of the source concept ``bear" so that the MSE loss is reduced.
\begin{equation}
\text{Loss}_{mse} = \frac{1}{d} \sum_{i=1}^d \left( h_{\text{orig}}, h_{\text{new}}\right)^2
\label{eq:mse_loss}
\end{equation}
The distance between each $\mathrm{wte}_{\text{orig}}$ and $\mathrm{wte}_{\text{new}}$ varies and depends on the chosen pre-train models. Consequently, achieving optimal edit performance requires different numbers of optimization steps for different tokens. The stopping threshold $\tau$ in line 4 in Algorithm~\ref{alg:pseduo_code} adjusts these optimization steps, where $\lambda \in [0, 1]$ denotes the optimization strength to reduce the distance between $\mathrm{wte}_\text{orig}$ and $\mathrm{wte}_\text{new}$ to a fraction of its initial value. Empirically, we found $\lambda$ straightforward to tune, as the value optimized on one or two examples generalizes well to other cases, and we observe that 0.2 or 0.3 works effectively across all instances of TIMED.

\begin{algorithm}[t!]
\small
\caption{\ee{} for a single edit} 
\label{alg:pseduo_code}
\KwIn{Stable diffusion model $\mathbf{M}$, 
Edit pair set $\mathbf{E} = \{p_{\text{orig}} , p_{\text{new}}\}$,
Maximum iterations $T$, Stopping ratio $\lambda$, Original token index $\mathrm{I}_\text{orig}$
}
\KwResult{Modified token embedding for $\mathbf{M}$ 
}
Initialize optimizer for source word tokens $\mathrm{wte}_{\text{orig}} = \mathbf{M}.\text{text\_encoder.WTE.weight}[ \mathrm{I}_\text{orig} ]$;
Initialize MSE loss function $\mathcal{L}_{\text{MSE}}$\;
    \ForEach{$(p_{\text{orig}}, p_{\text{new}})$ in $E$}{
    Precompute initial last hidden state $\mathbf{h}_{\text{orig}}^{\text{init}}$ and $\mathbf{h}_{\text{new}}$\;
    Precompute stop threshold $\tau=\lambda \cdot \mathcal{L}_{\text{MSE}}(h_{\text{orig}}^{\text{init}}, h_{\text{new}}) $\;
    
    \For{$i = 1$ to $T$}{
        Compute updated last hidden state $\mathbf{h}_{\text{orig}}^{i}$\;
        Calculate Loss $\mathcal{L} = \mathcal{L}_{\text{MSE}}(\mathbf{h}_{\text{orig}}^i, \mathbf{h}_{\text{new}})$\;    
        
        \If{$\mathcal{L} \leq \tau$}{
            \textbf{break} \# stop optimization\;
        }
        Update original WTE via $\mathcal{L}$.step\;
    }
        Update model $\mathbf{M}$ with optimized $\mathrm{wte}_{\text{orig}}$\;
}
\Return Edit token embedding for $\mathbf{M}$
\label{eq:pseduo_code}
\end{algorithm}

\begin{table}[h!]
\centering
\resizebox{.9\columnwidth}{!}{
\begin{tabular}{l|cc|cc}
\hline
& \multicolumn{2}{c|}{TIME} & \multicolumn{2}{c}{\ee{}} \\
\hline
& SD 1.4 & SD XL & SD 1.4 & SD XL \\
\hline
FLOP & 19,169,280 & 340,787,200 & \textbf{1,536} & \textbf{4,096} \\
Weight & 2.200\% & 9.625\% & \textbf{0.002}\% & \textbf{0.003}\% \\

\hline
\end{tabular}
}
\caption{A comparison between TIME~\cite{orgad2023editing} and \ee{} based on the average FLOPs for each edit and the ratio of edited weights required to modify a single object in the model.
}
\label{fig:FLOP}
\end{table}

Our method is exceptionally parameter-efficient: we only tune one token's embedding, a vector of 768 dimensions.\footnote{In some cases, we tune multiple tokens' WTEs when the object word is split into subwords of multiple tokens or consists of multiple words.} This accounts for merely 0.002\% of the 
total parameter of Stable Diffusion 1.4 and 0.003\% of Stable Diffusion XL, 1000 times less than TIME~\citep{orgad2023editing}, with 2.20\% and 9.62\% respectively. 

%% file: sec/4_eval.tex
\section{Edit Implicit Assumption}\label{sec:experiment}

\subsection{Implementation details}\label{sec:model_and_implementation}

\begin{figure}[hb!]
    \centering
    \includegraphics[width=1\linewidth]{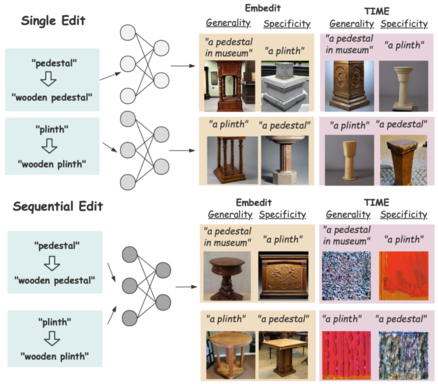}
    \caption{Illustration of single editing and sequential editing modes. In single editing mode, one model is edited once for one object, so a model for ``pedestal'' and a separate model for ``plinth''. 
    In sequential editing, one model is edited for both ``pedestal'' and ``plinth''. 
    }
    \label{fig:compare_single_seq}
\end{figure}

We compare \ee{} with TIME~\citep{orgad2023editing} in two editing modes: single edit (one edit one model which is the setup in TIME~\citep{orgad2023editing}, illustrated on the top of Figure~\ref{fig:compare_single_seq}) and sequential edit (one model is edited for multiple objects, the bottom of Figure~\ref{fig:compare_single_seq}), across two model sizes: Stable Diffusion v1.4 (SD 1.4) and Stable Diffusion XL (SD XL). We use the same SD 1.4 \footnote{\url{https://huggingface.co/CompVis/stable-diffusion-v1-4}} on Huggingfaceas TIME~\citep{orgad2023editing} and the official SD XL \footnote{\url{https://huggingface.co/stabilityai/stable-diffusion-xl-refiner-1.0}} from Huggingface as the backbone text-to-image model.

In our implementation of TIME~\citep{orgad2023editing}, we adopt their suggested default hyperparameters for SD 1.4. However, we discover that applying the default $\lambda$ to SD XL leads to complete editing failure. To ensure fair comparison, for SD XL, we tune the hyperparameter $\lambda$ by grid search between 0.01 and 3,000 (TIME's default for single editing is 0.1), and find that $\lambda=50$ works best. 

All experiments are run on an NVIDIA A100, using a fixed random seed for consistency. 
Further implementation details are provided in the supplementary.

\subsection{Dataset}
For single editing, following~\citep{orgad2023editing}, we use TIMED, a dataset~\citep{orgad2023editing} of 104 entries, as the text-to-image model editing dataset. Each entry contains an edit pair of an original object and a new object, used for one embedding editing. See Table~\ref{tab:editing_testing} for a sample entry. The original object (e.g., ``bear'') is a generic token that describes a scenario where a visual attribute is implicitly inferred by the model. The new object (e.g., ``polar bear'') is more specific and describes the same scenario with a precise attribute. 

\begin{table}[h!]
\centering
    \resizebox{0.85\columnwidth}{!}{%
    \begin{tabular}{p{0.95cm}|p{2.4cm}|p{3.2cm}}
        \hline
        \textbf{Edit} & \textbf{Original} & \textbf{Destination} \\
        \hline
        & bear & polar bear \\
        \hline
        \textbf{Test} & \textbf{Original} & \textbf{Destination} \\
        \hline
        \multirow{5}{*}{\rotatebox{45}{\textbf{Positives}}} 
        & a zoo with bear & a zoo with polar bear \\
        & a bear on beach & a polar bear on beach \\
        & bear on the tree & polar bear on the tree\\
        & cubist bear  & cubist polar bear \\
        & little bear & little polar bear\\
        \hline
        \multirow{5}{*}{\rotatebox{45}{\textbf{Negatives}}} 
        & a panda & a polar panda \\
        & a dog & a polar dog \\
        & a cat & a polar cat\\
        & a koala & a polar koala \\
        & a sloth & a polar sloth\\
        \hline
    \end{tabular}
    }
    \caption{An example of a single edit in \ee{}}
    \label{tab:editing_testing}
\end{table}

For sequential editing, we remove the objects that are negatives for specificity evaluation in the single editing mode but positives in the sequential editing mode. For example, as shown in Figure~\ref{fig:compare_single_seq}, in the single editing, when evaluating the specificity for editing ``plinth'' to ``wooden plinth'' (not affecting unrelated non-target objects), we consider it fails if the model outputs ``wooden pedestal'' when prompts ``pedestal'' as ``pedestal'' is not the edit target. 
However, we consider it as an edit success in sequential editing where the target is to edit both ``pedestal'' to ``wooden pedestal'' and ``plinth'' to ``wooden plinth'' in the same model. As a result, we have 77 entries for the sequential editing. Further details are included in the supplementary materials.

\subsection{Evaluation}

\begin{figure*}[ht!]
    \centering
    \vspace{0.1cm}
    \textbf{Edit ``dog" to ``schnauzer dog"}
    \vspace{0.1cm}
    \\
    \begin{subfigure}[b]{0.34\textwidth}  
        \centering
        \includegraphics[width=0.3\linewidth]{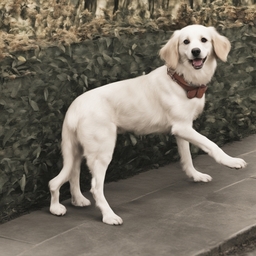}
        \hspace{-5pt} 
        \begin{overpic}[width=0.3\linewidth]{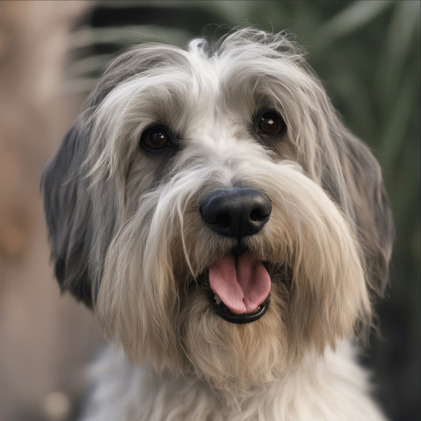}
            \put(-30, 25){\includegraphics[width=0.15\linewidth]{figs/right_arrow_1.png}}
        \end{overpic}
        \hspace{-3pt}
        \includegraphics[width=0.3\linewidth]{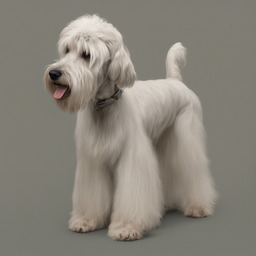}
        \caption{Efficacy: ``a dog''}
    \end{subfigure}
    \hspace{-10pt}
    \begin{subfigure}[b]{0.34\textwidth}  
        \centering
        \includegraphics[width=0.3\linewidth]{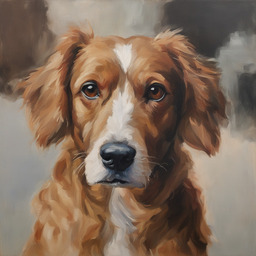}
        \hspace{-5pt}
        \begin{overpic}[width=0.3\linewidth]{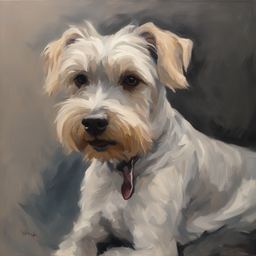}
            \put(-30, 25){\includegraphics[width=0.15\linewidth]{figs/right_arrow_1.png}}
        \end{overpic}
        \hspace{-3pt}
        \includegraphics[width=0.3\linewidth]{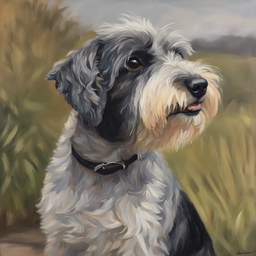}
        \caption{Generality: ``an oil painting of a dog''}
    \end{subfigure}
    \hspace{-10pt}
    \begin{subfigure}[b]{0.34\textwidth}  
        \centering
        \includegraphics[width=0.3\linewidth]{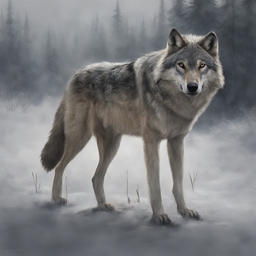}
        \hspace{-5pt}
        \begin{overpic}[width=0.3\linewidth]{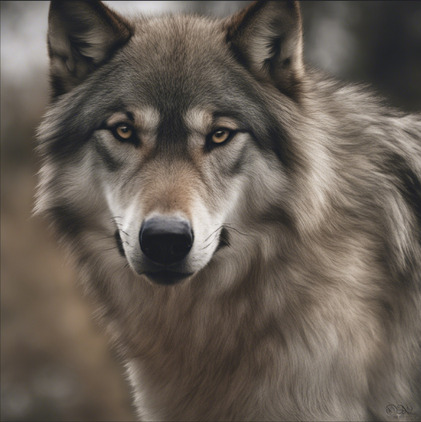}
            \put(-30, 25){\includegraphics[width=0.15\linewidth]{figs/right_arrow.png}}
        \end{overpic}
        \hspace{-3pt}
        \includegraphics[width=0.3\linewidth]{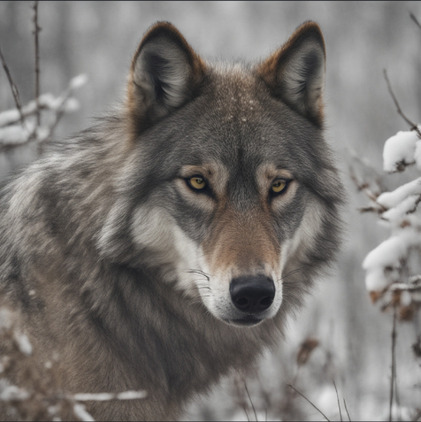}
        \caption{Specificity: ``a wolf''}
    \end{subfigure}

    \caption{Illustration of Efficacy, Generality, and Specificity. Images are generated by \ee{}-edited Stable Diffusion XL.}
    \label{fig:Efficacy_G_S}
\end{figure*}

Following TIME~\cite{orgad2023editing}, We assess edit performance using \emph{efficacy}, \emph{generality}, and \emph{specificity} metrics, evaluated with the CLIP ViT-B/32 model~\cite{radford2021learning} as a zero-shot text-based classifier. \textbf{Efficacy} measures the effectiveness of the editing method on the source prompt (see Figure~\ref{fig:Efficacy_G_S} (a)). \textbf{Generality} assesses the method's adaptability to similar prompts, tested using the positive prompts (Figure~\ref{fig:Efficacy_G_S} (b)). \textbf{Specificity} evaluates the method's precision in avoiding unintended changes, tested with negative prompts (Figure~\ref{fig:Efficacy_G_S} (c)). As~\cite{rombach2022high,saharia2022photorealistic,ramesh2022hierarchical}, we evaluate the image generation performance of edited models using FID~\cite{heusel2017gans}, CLIP Score~\cite{hessel2021clipscore}.
\textbf{FID}~\cite{heusel2017gans} assesses image quality by measuring similarity to the MS-COCO validation set (resized to 512×512)~\cite{lin2014microsoft}. \textbf{CLIP Score} evaluates text-image alignment, ensuring content matches the descriptions. We randomly sample 3,000 captions from the MS-COCO dataset to test the effect of modifications.

We report the results for both the baseline and oracle settings as TIME~\cite{orgad2023editing}. The baseline represents the unedited model's performance using only the source prompt for all image generations. The oracle, on the other hand, uses the same undefined model with destination positive prompts for positive samples and source negative prompts for negative samples. The oracle serves as an upper bound for the potential performance achievable by text-based model editing techniques.

\subsection{Results}\label{sec:object_results}

\begin{table*}[b]
\centering
\resizebox{\textwidth}{!}{%
\begin{tabular}{@{}lcc|cc|cc|cc|cc|cc@{}}
\toprule
\multicolumn{1}{c}{\multirow{2}{*}{}} & \multicolumn{2}{c|}{SD 1.4} & \multicolumn{2}{c|}{SD XL} & \multicolumn{2}{c|}{SD 1.4 (single edit)} & \multicolumn{2}{c|}{SD 1.4 (sequential edit)} & \multicolumn{2}{c|}{SD XL (single edit)} & \multicolumn{2}{c}{SD XL (sequential edit)} \\ \cline{2-13} 
\multicolumn{1}{c}{} & Oracle & Baseline & Oracle & Baseline & TIME & EmbEdit & TIME & EmbEdit & TIME (XL) & EmbEdit(XL) & TIME (XL) & EmbEdit(XL) \\ \hline
Efficacy ($\uparrow$) & 98.7 & 11.04 & 97.7 & 8.67 & 87.17 & \textbf{93.18} & NaN & \textbf{96.59} & 81.01 & \textbf{92.86} & NaN & \textbf{90.58} \\
 & \footnotesize±0.64 & \footnotesize±2.64 & \footnotesize±0.99 & \footnotesize±2.28 & \footnotesize±2.62 & \footnotesize±2.39 &  & \footnotesize±1.44 & \footnotesize±3.17 & \footnotesize±1.99 &  & \footnotesize±2.61 \\
Generality ($\uparrow$) & 94.71 & 12.01 & 95.23 & 8.23 & 69.93 & \textbf{82.74} & NaN & \textbf{86.36} & 51.43 & \textbf{77.86} & NaN & \textbf{89.19} \\
 & \footnotesize±0.78 & \footnotesize±1.59 & \footnotesize±0.86 & \footnotesize±1.38 & \footnotesize±2.72 & \footnotesize±2.51 &  & \footnotesize±1.99 & \footnotesize±3.15 & \footnotesize±3.04 & & \footnotesize±3.00 \\
Specificity ($\uparrow$) & 88.57 & 88.56 & 94.28 & 94.2 & 66.49 & \textbf{77.09} & NaN & \textbf{69.92} & 79.35 & \textbf{87.71} & NaN & \textbf{76.92} \\
 & \footnotesize±1.82 & \footnotesize±1.82 & \footnotesize±0.97 & \footnotesize±0.98 & \footnotesize±2.28 & \footnotesize±2.33 &  & \footnotesize±2.55 & \footnotesize±2.3 & \footnotesize±1.57 &  & \footnotesize±2.02 \\ \hline
FID ($\downarrow$) & 40.13 & 40.13 & 37.65 & 37.65 & 40.46 & 40.71 & 243.04 & 40.12 & 37.97 & 37.26 & 308.06 & 38.18 \\
CLIP Score ($\uparrow$) & 31.17 & 31.17 & 31.66 & 31.66 & 31.19 & 31.15 & 30.75 & 18.92 & 31.66 & 31.70 & 21.51 & 31.41 \\ \hline
\end{tabular}
}
\caption{Edit performance and generative quality comparison on SD 1.4 and SD XL. \% is omitted for clarity. \textbf{Best} for each model, metrics, and editing mode is highlighted \textbf{in bold} (oracle is excluded). The standard deviation is shown below. NaN: sequential editing with TIME causes Stable Diffusion to collapse and only generate salt-and-pepper noise images, as shown in Figure~\ref{fig:embedit_illustrate}.}
\label{tab:object_edit_results_xl_1.4}
\end{table*}

Overall, as shown in Table~\ref{tab:object_edit_results_xl_1.4}, \ee{} consistently outperforms TIME across the three metrics, the two models in both editing modes by a substantial margin. For example, in the single editing, the greatest performance gap is observed for Generality on SD XL, where \ee{} achieves 77.86\% compared to TIME's 51.43\%, and the narrowest gap appears in Efficacy on SD 1.4, with \ee{} reaching 93.18\% and TIME achieving 87.17\%. In terms of generation quality, measured by FID and CLIP Score, \ee{} maintains comparable performance to the unedited model (upper bound baseline) across both editing modes and both models. In contrast, TIME performs poorly in the sequential mode, as evidenced by its significantly degraded FID scores of 243.04 for SD 1.4 and 308.06 for SD XL.

\begin{figure}[hb!]
    \centering
    \begin{minipage}{0.4222\textwidth}
        \centering
        \resizebox{\columnwidth}{!}{%
            \begin{tikzpicture}
            \begin{axis}[
                width=10cm, 
                height=7cm, 
                xlabel={FLOP}, xlabel style={at={(1,0)},anchor=west,},
                ylabel={Metrics (\%)},
                xmode=log,
                xmin=1000, xmax=500000000,
                ymin=40, ymax=100,
                xtick={1536,6000, 19169280,340787200},
                xticklabels={1536, 4096, $1.9 \times 10^7$, $3.4 \times 10^8$},
                title={Single edit},
                xticklabel style={
                /pgf/number format/1000 sep=,
                font=\small,
                rotate=10,
                anchor=north east,},
                legend style={at={(0.5,-0.3)}, anchor=north, legend columns=3, /tikz/every even column/.append style={column sep=0.5cm}, font=\small}
            ]
                \addplot[only marks, mark=*, color=blue] 
                    coordinates {(340787200, 81.01)} 
                    node[anchor=south east, scale=.9, ] {TIME XL};
                \addplot[only marks, mark=square*, color=blue] 
                    coordinates {(340787200, 51.43)}
                    node[anchor=east, scale=.9, ] {TIME XL};
                \addplot[only marks, mark=triangle*, color=blue] 
                    coordinates {(340787200, 79.35)}
                    node[anchor=north east, scale=.9, ] {TIME XL};

                \addplot[only marks, mark=*, color=cyan] 
                    coordinates {(19169280, 87.17)}
                    node[anchor=south, scale=.9, ] {TIME v1-4};
                \addplot[only marks, mark=square*, color=cyan] 
                    coordinates {(19169280, 69.93)}
                    node[anchor=south, scale=.9, ] {TIME v1-4};
                \addplot[only marks, mark=triangle*, color=cyan] 
                    coordinates {(19169280, 66.49)}
                    node[anchor=east, scale=.9, ] {TIME v1-4};
    
                \addplot[only marks, mark=*, color=orange] 
                    coordinates {(1536, 93.18)}
                    node[anchor=south west, scale=.9, ] {\ee{} v1-4};
                \addplot[only marks, mark=square*, color=orange] 
                    coordinates {(1536, 82.74)}
                    node[anchor=west, scale=.9, ] {\ee{} v1-4};
                \addplot[only marks, mark=triangle*, color=orange] 
                    coordinates {(1536, 77.09)}
                    node[anchor=south west, scale=.9, ] {\ee{} v1-4};
                
                
                \addplot[only marks, mark=*, color=red] 
                    coordinates {(4096, 92.86)}
                    node[anchor=west, scale=.9, ] {\ee{} XL};
                \addplot[only marks, mark=square*, color=red] 
                    coordinates {(4096, 77.86)}
                    node[anchor=north west, scale=.9, ] {\ee{} XL};
                \addplot[only marks, mark=triangle*, color=red] 
                    coordinates {(4096, 87.71)} 
                    node[anchor=west, scale=.9, ] {\ee{} XL};
            \end{axis}
        \end{tikzpicture}
        }
    \end{minipage}
        \begin{minipage}{0.4222\textwidth}
        \centering
        \resizebox{\columnwidth}{!}{%
            \begin{tikzpicture}
            \begin{axis}[
                width=10cm, 
                height=7cm, 
                xlabel={FLOP}, xlabel style={at={(1,0)},anchor=west,},
                ylabel={Metrics (\%)},
                xmode=log,
                xmin=1000, xmax=500000000,
                ymin=0, ymax=100,
                xtick={1536,6000, 19169280,340787200},
                xticklabels={1536, 4096, $1.9 \times 10^7$, $3.4 \times 10^8$},
                title={Sequential edit},
                xticklabel style={
                /pgf/number format/1000 sep=,
                font=\small,
                rotate=5,
                anchor=north east,},
                legend style={at={(0.5,-0.15)}, anchor=north, legend columns=3, /tikz/every even column/.append style={column sep=0.5cm}, font=\small}]
                \addlegendimage{only marks, mark=*, color=black,line width=2}
                \addlegendentry{Efficacy}
                \addlegendimage{only marks, mark=square*, color=black,line width=2}
                \addlegendentry{Generality}
                \addlegendimage{only marks, mark=triangle*, color=black,line width=2}
                \addlegendentry{Specificity}
                
                \addplot[only marks, mark=diamond*, color=blue] 
    coordinates {(340787200, 0)} 
    node[anchor=south east, scale=0.9] {TIME XL};

\addplot[only marks, mark=diamond*, color=cyan] 
    coordinates {(19169280, 0)} 
    node[anchor=south east, scale=1] {TIME v1-4};

                \addplot[only marks, mark=*, color=orange] 
                    coordinates {(1536, 96.59)}
                    node[anchor=west, scale=.9, ] {\ee{} v1-4};
                
                \addplot[only marks, mark=square*, color=orange] 
                    coordinates {(1536, 86.36)}
                    node[anchor=north west, scale=.9, ] {\ee{} v1-4};
                
                \addplot[only marks, mark=triangle*, color=orange] 
                    coordinates {(1536, 69.92)}
                    node[anchor=north west, scale=.9, ] {\ee{} v1-4};
                
                \addplot[only marks, mark=*, color=red] 
                    coordinates {(4096, 90.58)}
                    node[anchor=west, scale=.9, ] {\ee{} XL};
                
                \addplot[only marks, mark=square*, color=red] 
                    coordinates {(4096, 89.19)}
                    node[anchor=north west, scale=.9, ] {\ee{} XL};
                
                \addplot[only marks, mark=triangle*, color=red] 
                    coordinates {(4096, 76.92)} 
                    node[anchor=north west, scale=.9, ] {\ee{} XL};
            \end{axis}
        \end{tikzpicture}
            }
    \end{minipage}

    \caption{A comparison of Efficacy, Generality, Specificity, and FLOP between \ee{} and TIME. The closer to the right top corner, the better. Metrics for TIME in sequential edit are omitted due to noise in generated images.}
    \label{fig:performance_comparison}
\end{figure}
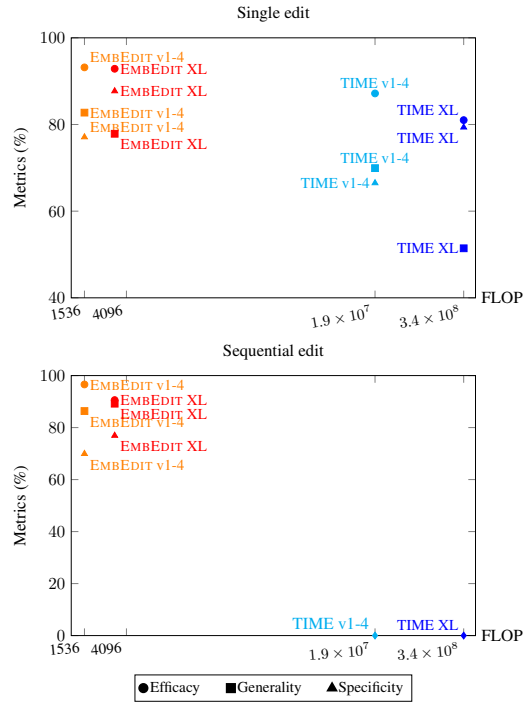

We compare editing performance and computational efficiency in Figure~\ref{fig:performance_comparison}. The top-right corner represents the optimal scenario, where high editing performance is achieved with minimal computation. \ee{} (in red and orange) demonstrate superior performance, achieving higher editing accuracy with significantly lower FLOPs across various metrics and model sizes.

\paragraph{Sequential editing} \ee{} maintains robust performance in the sequential editing. For SD 1.4 specifically, considering the standard deviation, sequential editing does not affect the edit performance, achieving 93.18\% (2.39\%) and 96.59\% (1.44\%) efficacy for single and sequential editing, respectively. On the other hand, the sequential editing by TIME leads both models to collapse, only outputing salt-and-pepper noise, as shown in the red-bordered images in Figure~\ref{fig:compare_single_seq}. 

\paragraph{Generalization on Stable Diffusion XL} \ee{} demonstrates comparable editing performance and generation quality across models of different sizes. Interestingly, the larger model size leads an increase in Specificity (77.09\% $\rightarrow$ 87.71\% in single editing and 69.92\% $\rightarrow$ 76.92\% in sequential editing). We also observe this pattern with TIME that specificity increases from 66.49\% to 79.35\% in single editing mode. One potential explanation is that the dual text encoders in SDXL helps constrain semantic changes locally through the model.

\begin{figure}[t]
    \centering
    \includegraphics[width=1\linewidth]{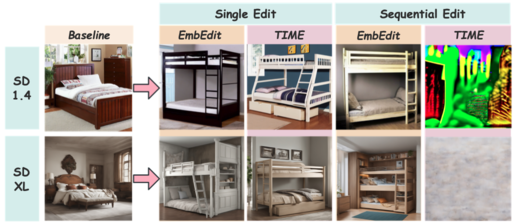}
    \caption{A comparison of edit performance between \ee{} and TIME methods in Stable Diffusion v1.4 and Stable Diffusion XL models.
    }
    \label{fig:comparison_beds}
\end{figure}

\begin{figure}[hb!]
    \centering
    \vspace{0.1cm}
    {\small \textbf{Edit ``ice cream" to ``strawberry ice cream"}}
    \vspace{0.1cm}
        \resizebox{0.475\textwidth}{!}{%
        \begin{subfigure}[b]{0.3\textwidth}  
            \centering
            \includegraphics[width=0.3\linewidth]{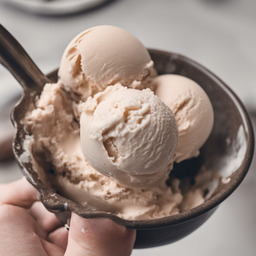}
            \hspace{-5pt} 
            \begin{overpic}[width=0.3\linewidth]{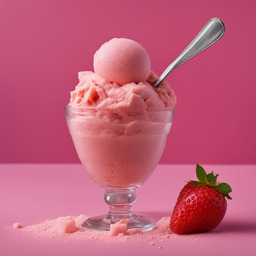}
                \put(-35, 20){\includegraphics[width=0.2\linewidth]{figs/right_arrow_1.png}}
            \end{overpic}
            \hspace{-3pt}
            \includegraphics[width=0.3\linewidth]{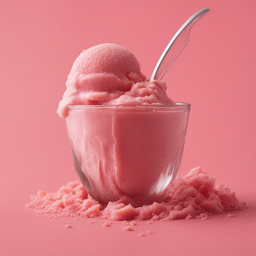}
            \caption*{``a scoop of ice cream''}
        \end{subfigure}

        \begin{subfigure}[b]{0.3\textwidth}
            \centering
            \includegraphics[width=0.3\linewidth]{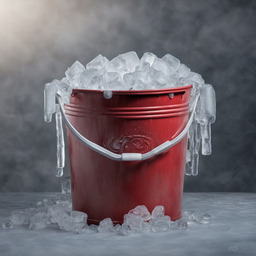}
            \hspace{-5pt}
            \begin{overpic}[width=0.3\linewidth]{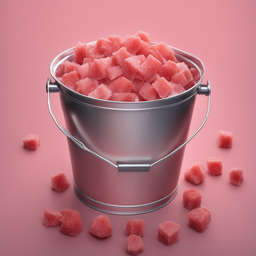}
                \put(-35, 20){\includegraphics[width=0.2\linewidth]{figs/right_arrow.png}}
            \end{overpic}
            \hspace{-3pt}
            \includegraphics[width=0.3\linewidth]{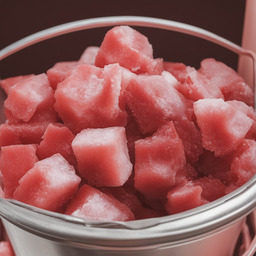}
            \caption*{``a bucket of ice''}
        \end{subfigure}
    }
    \\
    {\small \textbf{Edit ``mushroom'' to ``purple mushroom''}}
    \vspace{0.1cm}
    \resizebox{0.475\textwidth}{!}{%
        \begin{subfigure}[b]{0.3\textwidth}  
            \centering
            \includegraphics[width=0.3\linewidth]{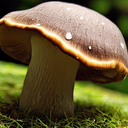}
            \hspace{-5pt} 
            \begin{overpic}[width=0.3\linewidth]{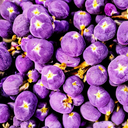}
                \put(-35, 20){\includegraphics[width=0.2\linewidth]{figs/right_arrow_1.png}}
            \end{overpic}
            \hspace{-3pt}
            \includegraphics[width=0.3\linewidth]{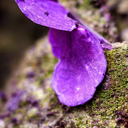}
            \caption*{``a mushroom''}
        \end{subfigure}

        \begin{subfigure}[b]{0.3\textwidth}
            \centering
            \includegraphics[width=0.3\linewidth]{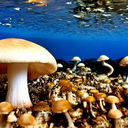}
            \hspace{-5pt}
            \begin{overpic}[width=0.3\linewidth]{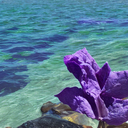}
                \put(-35, 20){\includegraphics[width=0.2\linewidth]{figs/right_arrow.png}}
            \end{overpic}
            \hspace{-3pt}
            \includegraphics[width=0.3\linewidth]{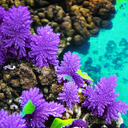}
            \caption*{``mushroom in the sea''}
        \end{subfigure}
    }
    \\
    \caption{Illustration of failure cases where an object consists of multiple tokens (top) and where the target concept rarely appears in daily life.}
    \label{fig:limitation_examples}
\end{figure}
\paragraph{Qualitative Analysis on Failures} We investigate failure cases and identify two distinct patterns. First, as illustrated in the top row of Figure~\ref{fig:limitation_examples}, \ee{} fails specificity when editing multi-word objects with broad semantic meanings. A prime example is ``ice cream'', where both constituent words are common umbrella terms with multiple semantic interpretations in English. When editing ``ice cream" to ``strawberry ice cream", \ee{} jointly edit ``ice'' and ``cream'' to fuse the assumption ``strawberry'' to their WTEs. The successful editing causes the word ``ice'' to inappropriately carry the ``strawberry'' attribute of red color even in unrelated contexts, such as ``a bucket of ice''. This issue becomes particularly problematic when the constituent words frequently appear in semantically unrelated compound terms, such as ``ice hockey'' and ``face cream''. From a linguistic perspective, these components function as hypernyms or superordinates, encompassing broader semantic categories~\cite{10.1145/3501714.3501736}. Consequently, editing these terms risks unintended modifications to semantically distinct expressions that contain the edited words like ``ice hockey''.

Second, the effectiveness of \ee{} diminishes for concepts with limited real-world occurrence. As illustrated in the second row of Figure~\ref{fig:limitation_examples}, the generated objects fail to preserve the characteristic mushroom morphology as ``purple mushroom'' rarely appears. This observation aligns with previous hypotheses suggesting that diffusion generators encode perceptual attributes of objects~\cite{basu2024localizing}.



%% file: sec/5_gender.tex
\section{Gender Bias Mitigation}\label{sec:gender}



In the previous section, we introduced \ee{} for editing implicit model assumptions. This section addresses social bias as a specific type of implicit assumption encoded within language models~\cite{Blodgett2020LanguageI,Devinney2022TheoriesO,may2019measuring} and text-to-image diffusion models~\cite{Fraser2023AFF,Struppek2022TheBA,Zameshina2023DiverseDE, arad-etal-2024-refact,Xiong2024EditingMC,Masrourisaadat2024AnalyzingQB,Mandal2023MultimodalBA}. They are known to inherently reflect social and cultural biases~\cite{bender2021dangers}. For example, Stable Diffusion v1.4 often associates specific genders with professions: only 5.55\% of images generated for ``A photo of a CEO'' depict women, while 97.22\% of images for ``A photo of a housekeeper'' feature women. 
Our goal is to mitigate these stereotype-driven assumptions in Stable Diffusion.\footnote{We limit our analysis to binary genders to avoid misrepresenting non-binary identities. Future work should thoughtfully expand to include the full gender spectrum.
}

The original Stable Diffusion model exhibits varying levels of bias across professions. For instance, editing a profession like ``CEO'' requires a lower \( \lambda \)  to achieve a stronger bias reduction, while a profession like ``teacher'', which has less inherent bias, requires a high \( \lambda \). As practice in \cite{orgad2023editing}, we manually set a lower \( \lambda \) for professions with stronger biases and a slightly higher \( \lambda \) for those with less bias.

\subsection{Dataset}

We experiments with six professions as \cite{orgad2023editing} with under-specified source prompts in the form ``A/An [profession]'', such as ``A CEO''. The destination prompt specifies a non-stereotypical gender, such as ``A female CEO''. We add five test prompts for each profession, describing it in various scenarios, e.g., ``A CEO laughing'', as illustrated in Table~\ref{tab:editing_gender}

\begin{table}[h]
\small
\centering
\begin{tabular}{l|l|l}
\hline
\multirow{1}{*}{}& Source & Destination \\
\hline
\multirow{1}{*}{Editing} & CEO & male CEO \\
\hline
\multirow{1}{*}{Validation} & \multicolumn{2}{l}{A photo of a CEO} \\
\hline
\multirow{5}{*}{Testing} & \multicolumn{2}{l}{A painting of a CEO} \\
& \multicolumn{2}{l}{A CEO working} \\
& \multicolumn{2}{l}{A CEO laughing} \\
& \multicolumn{2}{l}{A CEO in the workplace} \\
& \multicolumn{2}{l}{A CEO digital art} \\
\hline
\end{tabular}
\caption{An example entry in mitigating gender bias dataset.}
\label{tab:editing_gender}
%
\end{table}

\subsection{Evaluation} 
For each profession \( p \), we aim for 50\% of the generated images to depict women and 50\% men. 
To assess the level of gender bias, we calculate the percentage of female figures generated for each profession, denoted as \( F_p \in [0, 100] \)~\cite{orgad2023editing}. We generate 24 images for each test prompt and use CLIP\footnote{For gender result, we use CLIP ViT-B/32 model to to evaluate the gender tendency of generated images. Different from previous efficacy test, this test focus on determining whether the generated images are more similar to ``[female] [profession]'' or``[male] [profession]''. Additionally, we conduct a human evaluation on 100 examples, the results show that this method is 100\% accuracy in distinguishing the gender tendency of the generated images.} to classify gender in each image. Therefore, for each profession, we generate 144 (6*24) images to calculate the percentage of females. The optimal value for \( F_p \) is 50, indicating an equal representation of male and female images. 
We compare the editing performance \( F_p \) to the baseline unedited model. The oracle is defined as the unedited model explicitly asked with ``a [gender] [profession]'', where [gender] is randomly set to ``female'' or ``male'' in each generation.



\subsection{Results}


\begin{table}[ht!]
\centering

\resizebox{\columnwidth}{!}{%
\begin{tabular}{cl|cc|cc}
\hline
& & \multicolumn{1}{c}{Baseline} & \multicolumn{1}{c}{Oracle} & \multicolumn{1}{c}{TIME} & \multicolumn{1}{c}{\ee{}} \\
\hline
\multirow{6}{*}{\( F_p \)} & Hairdresser & 77.08 & 53.47 & 47.50 & \textbf{49.30} \\
& CEO & 5.55 & 55.56 & 33.33 & \textbf{39.58}\\
& Teacher & 80.55 & 48.61 & 24.17 & \textbf{57.63} \\
& Lawyer & 29.86 & 44.45 & 59.17 & \textbf{55.84} \\
& Housekeeper & 97.22 & 57.64 & 86.67 & \textbf{43.75} \\
& Farmer & 3.47 & 55.56 & 48.43 & \textbf{55.56} \\
\hline
$\Delta (\downarrow)$ & & 0.598 & 0.097 & 0.308 & \textbf{0.121}  \\
\hline
\end{tabular}%
}
\caption{Results of mitigating gender bias in profession assumptions. Percentages (\%) are omitted for clarity. For $F_p$, ``{50}" represents the ideal debiased result (50\% female, 50\% male). The best result in each row is highlighted in bold. $\Delta (\downarrow)$ indicates the average deviation from 50, with smaller values reflecting a more neutral gender assumption.}
\label{tab:gender_edit_results}
\end{table}

As shown in Table~\ref{tab:gender_edit_results}, \ee{} consistently outperforms TIME across all professional categories, reducing the overall $\Delta$ from 0.598 to 0.121, which is substantially more effective than TIME's reduction to 0.308. For instance, \ee{} reduces bias within ``teacher'' from 80.55\% to 57.63\%, compared to TIME's 24.17\%. Figure~\ref{fig:gender_lawyer} demonstrates the bias mitigation performance for ``teacher'' and ``lawyer''. 

We further explore an automatic method that uses a different loss function and stopping criterion from Section~\ref{sec:experiment} to adjust gender bias during editing, i.e. adjusting the distance between ``[profession]'' and both ``[male] [profession]'' and ``[female] [profession]''. The idea is to balance the representation distance from ``[profession]'' to ``[male] [profession]'' and from ``[profession]'' to ``[female][profession]''. As anticipated, the automatic method does not outperform manually adjusted settings. Further details on the methods and results are provided in the supplementary materials. 

\begin{figure}[h]
    \centering
    \resizebox{0.47\textwidth}{!}{%
    \begin{subfigure}{0.5\textwidth}
        \centering
        \begin{subfigure}{0.31\textwidth}
            \centering
            \includegraphics[width=\linewidth]{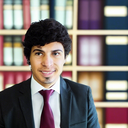}
        \end{subfigure}
        \begin{subfigure}{0.31\textwidth}
            \centering
            \includegraphics[width=\linewidth]{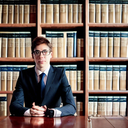}
        \end{subfigure}
        \begin{subfigure}{0.31\textwidth}
            \centering
            \includegraphics[width=\linewidth]{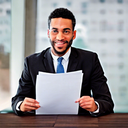}
        \end{subfigure}
        \begin{subfigure}{0.31\textwidth}
            \centering
            \includegraphics[width=\linewidth]{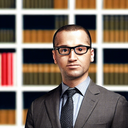}
        \end{subfigure}
        \begin{subfigure}{0.31\textwidth}
            \centering
            \includegraphics[width=\linewidth]{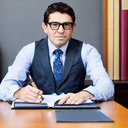}
        \end{subfigure}
        \begin{subfigure}{0.31\textwidth}
            \centering
            \includegraphics[width=\linewidth]{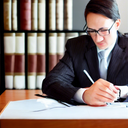}
        \end{subfigure}
        \caption*{\Large Baseline}
    \end{subfigure}  
    \hfill

 \begin{subfigure}{0.5\textwidth}
        \centering
        \begin{subfigure}{0.31\textwidth}
            \centering
            \includegraphics[width=\linewidth]{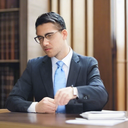}
        \end{subfigure}
        \begin{subfigure}{0.31\textwidth}
            \centering
            \includegraphics[width=\linewidth]{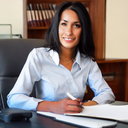}
        \end{subfigure}
        \begin{subfigure}{0.31\textwidth}
            \centering
            \includegraphics[width=\linewidth]{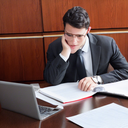}
        \end{subfigure}
        \begin{subfigure}{0.31\textwidth}
            \centering
            \includegraphics[width=\linewidth]{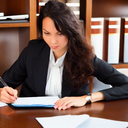}
        \end{subfigure}
        \begin{subfigure}{0.31\textwidth}
            \centering
            \includegraphics[width=\linewidth]{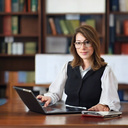}
        \end{subfigure}
        \begin{subfigure}{0.31\textwidth}
            \centering
            \includegraphics[width=\linewidth]{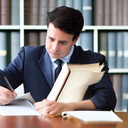}
        \end{subfigure}
        \caption*{\Large \ee{}}
    \end{subfigure}
    }
    \caption{Example of mitigating gender bias on \emph{lawyers}.}
    \label{fig:gender_lawyer}
\end{figure}

\begin{figure}[h]
\vspace{-.4cm}
    \centering
    \resizebox{0.47\textwidth}{!}{%
    \begin{subfigure}{0.5\textwidth}
        \centering
        \begin{subfigure}{0.31\textwidth}
            \centering
            \includegraphics[width=\linewidth]{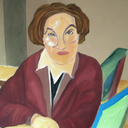}
        \end{subfigure}
        \begin{subfigure}{0.31\textwidth}
            \centering
            \includegraphics[width=\linewidth]{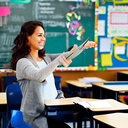}
        \end{subfigure}
        \begin{subfigure}{0.31\textwidth}
            \centering
            \includegraphics[width=\linewidth]{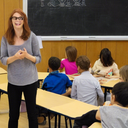}
        \end{subfigure}
        \begin{subfigure}{0.31\textwidth}
            \centering
            \includegraphics[width=\linewidth]{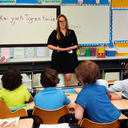}
        \end{subfigure}
        \begin{subfigure}{0.31\textwidth}
            \centering
            \includegraphics[width=\linewidth]{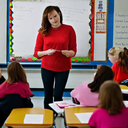}
        \end{subfigure}
        \begin{subfigure}{0.31\textwidth}
            \centering
            \includegraphics[width=\linewidth]{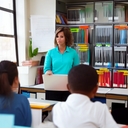}
        \end{subfigure}
        \caption*{\Large Baseline}
    \end{subfigure}  
    \hfill

 \begin{subfigure}{0.5\textwidth}
        \centering
        \begin{subfigure}{0.31\textwidth}
            \centering
            \includegraphics[width=\linewidth]{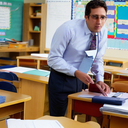}
        \end{subfigure}
        \begin{subfigure}{0.31\textwidth}
            \centering
            \includegraphics[width=\linewidth]{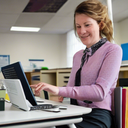}
        \end{subfigure}
        \begin{subfigure}{0.31\textwidth}
            \centering
            \includegraphics[width=\linewidth]{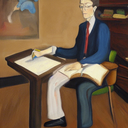}
        \end{subfigure}
        \begin{subfigure}{0.31\textwidth}
            \centering
            \includegraphics[width=\linewidth]{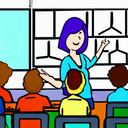}
        \end{subfigure}
        \begin{subfigure}{0.31\textwidth}
            \centering
            \includegraphics[width=\linewidth]{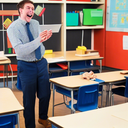}
        \end{subfigure}
        \begin{subfigure}{0.31\textwidth}
            \centering
            \includegraphics[width=\linewidth]{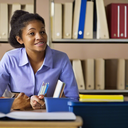}
        \end{subfigure}

        \caption*{\Large \ee{}}
    \end{subfigure}
    }
    \caption{Example of mitigating gender bias on \emph{teachers}.}
    \label{fig:gender_lawyer}
\end{figure}


%% file: sec/6_conclusion.tex
\section{Conclusions}

We present \ee{}, a simple yet effective approach for modifying implicit assumptions in text-to-image diffusion models by editing word token embeddings (WTEs). Our probing experiments provide intuitive motivation for this approach, showing that the WTE encodes sufficient information to represent visible attributes of objects. In experiments across two editing tasks,  \ee{} demonstrate state-of-the-art performance while being remarkably parameter-efficient, updating only 768 parameters (0.002\%) for Stable Diffusion v1.4 and 2048 parameters (0.003\%) for XL. Unlike previous methods, \ee{} maintains model stability during sequential editing and generalizes effectively across model scales.
Although \ee{} proves effective for editing implicit assumptions and mitigating gender bias, it shows limitations when handling multi-word objects. Future work addressing this could further enhance \ee{}'s capabilities. 

\section{Acknowledgements}
The computational resources for this project were provided by the Stantage HPC of the University of Sheffield. We also thank Zhenyang Liu for his invaluable assistance.



%% file: sec/X_suppl.tex
\clearpage
\setcounter{page}{1}
\renewcommand{\thepage}{S\arabic{page}}
\setcounter{figure}{0}
\setcounter{table}{0}

\twocolumn[
    \begin{center}
    {\Large \textbf{Supplementary Material}} 
    \addcontentsline{toc}{section}{Supplementary Material}
    \end{center}
]

This supplementary material provides additional information and analyses to complement the findings presented in the main manuscript. It is organized into three sections: The probing task is detailed in Sec.~\ref{sec:probe}, which contains the specific methodology and dataset preparation used to evaluate the WTE's performance in the color classification task; in Sec.~\ref{sec:exp_details}, additional experimental details, including datasets, model implementation, ablation study and an automatic gender method are provided; and Additional results, as shown in Sec.~\ref{sec:addition_results}, presents extended quantitative and qualitative evaluations with comparison to ReFACT. Together, these sections aim to enhance the reproducibility, transparency, and depth of our study.

\section{Probing Task}\label{sec:probe}

In the probing task, we prompt ChatGPT to generate two lists of objects: one comprising 100 red objects and another comprising 100 yellow objects. We divide the mix of the two lists into a training set and test set in random order, 80:20.
We extract the WTE (Word Token Embedding) of each object as their features. 
Then, we use these features to train a logistic regression model. During testing, the model predicts the color labels for the test set and achieves an accuracy of 90 (±1.25). 
This shows that WTE contains implicit assumptions.

\section{Additional Experimental Details}\label{sec:exp_details}

\subsection{Datasets}

\begin{figure}[ht]
    \centering
    \vspace{0.1cm}
    \textbf{\small Edit ``plinth" to ``wooden plinth"}
    \vspace{0.1cm}
    \\
    \begin{subfigure}[b]{0.45\textwidth}
    \begin{subfigure}{0.31\linewidth}
        \centering
        \includegraphics[width=0.9\linewidth]{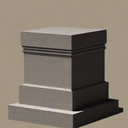}
        \caption*{\scriptsize Baseline``plinth''}
    \end{subfigure}
    \hfill
    \begin{subfigure}{0.31\linewidth}
        \centering
        \includegraphics[width=0.9\linewidth]{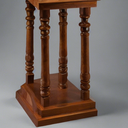}
        \caption*{\scriptsize \ee{} ``plinth''}
    \end{subfigure}
    \hfill
    \begin{subfigure}{0.31\linewidth}
        \centering
        \includegraphics[width=0.9\linewidth]{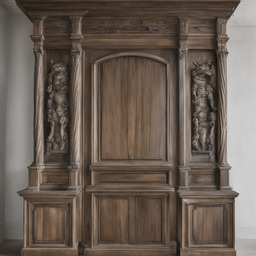}
        \caption*{\scriptsize Specificity ``pedestal''}
    \end{subfigure}
    \end{subfigure}
    \caption{A specificity test example for sequential edits: since ``pedestal'' is edited before ``plinth'', the ``plinth'' specificity test is considered a success.}
    \label{fig:sup_plinth}
\end{figure}

\begin{figure}[ht]
    \centering
    \vspace{0.1cm}
    \textbf{\small Edit ``dog" to ``Schnauzer dog"}
    \vspace{0.1cm}
    \\
    \begin{subfigure}[b]{0.45\textwidth}  
         \begin{subfigure}{0.31\linewidth}
        \centering
        \includegraphics[width=0.9\linewidth]{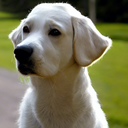}
        \caption*{\scriptsize Baseline ``dog''}
    \end{subfigure}
    \hfill
    \begin{subfigure}{0.31\linewidth}
        \centering
        \includegraphics[width=0.9\linewidth]{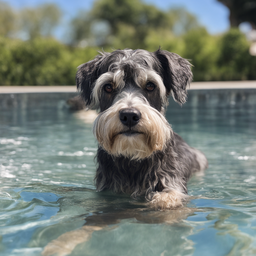}
        \caption*{\scriptsize \ee{} ``dog''}
    \end{subfigure}
    \hfill
    \begin{subfigure}{0.31\linewidth}
        \centering
        \includegraphics[width=0.9\linewidth]{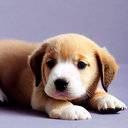}
        \caption*{\scriptsize Generality ``puppy''}
    \end{subfigure}
    \end{subfigure}
    \caption{An example of editing ``dog'' to ``schnauzer dog'': P2 is a successful edit, while P3 is a generality test with ``puppy''. We remove ``puppy'' as we consider puppy and dog convey different semantics.}
    \label{fig:sup_dog}
\end{figure}

\begin{figure}[h]
    \centering
    \vspace{0.1cm}
    \textbf{\small Edit ``subway" to ``subway pizza"}
    \vspace{0.1cm}
    \\
    \begin{subfigure}[b]{0.45\textwidth}
    \begin{subfigure}{0.31\linewidth}
        \centering
        \includegraphics[width=0.9\linewidth]{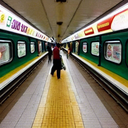}
        \caption*{\scriptsize Baseline``subway''}
    \end{subfigure}
    \hfill
    \begin{subfigure}{0.31\linewidth}
        \centering
        \includegraphics[width=0.9\linewidth]{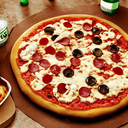}
        \caption*{\scriptsize \ee{} ``subway''}
    \end{subfigure}
    \hfill
    \begin{subfigure}{0.31\linewidth}
        \centering
        \includegraphics[width=0.9\linewidth]{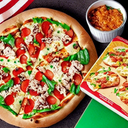}
        \caption*{\scriptsize ``a subway meal''}
    \end{subfigure}
    \end{subfigure}
    \caption{An example of ambiguous edit: ``subway" to ``subway pizza''. ``subway'' has dual meanings: food and transportation. Additionally, it is hard to determine whether the generated 
    images refer to a normal pizza or a ``subway pizza''.}
    \label{fig:sup_subway}
\end{figure}

The TIMED dataset reveals several limitations regarding sequential editing evaluation. Firstly and most notably, we observe instances where objects modified as the target in previous contexts appear in subsequent specificity test cases. Figure~\ref{fig:sup_plinth} gives an illustration of this.
Secondly, some generality test objects do not include the original objects. For example, the edit object is ``dog'' but the test prompt is ``puppy'', see Figure~\ref{fig:sup_dog} for details.
Thirdly, some instances are ambiguous and hard to evaluate. See Figure~\ref{fig:sup_subway} for details.
Also, the TIMED dataset contains several instances of impractical or surreal editing scenarios, which significantly compromise the model's performance.
For example, editing ``banana'' to ``blue banana'' introduces unnatural modifications that the model struggles to handle.
The list of those removed objects can be found in Table~\ref{tab:removed target objects}

\begin{table}[h!]
    \centering
    \small
    \begin{tabular}{|c|c|}
    \hline
    \multicolumn{1}{|c|}{\textbf{Old}} & \textbf{New} \\ \hline
    banana & blue banana \\
    cat & green cat \\
    dog & green dog \\
    fern & purple fern \\
    frog & purple frog \\
    panther & purple panther \\
    mushroom & purple mushroom \\
    pizza & square pizza \\
    root & purple root \\
    tree & purple tree \\
    Ron Weasley & female Ron Weasley \\
    Neville Longbottom & female Neville Longbottom \\
    truffle & purple truffle \\
    vehicle & flying vehicle \\
    Albus Dumbeldore & blond Albus Dumbeldore \\
    Draco Malfoy & female Draco Malfoy \\
    Hagrid & female Hagrid \\
    Harry Potter & female Harry Potter \\
    the sun & the green sun \\
    sunflower & blue sunflower \\
    McDonald's & McDonald's sushi \\
    subway & subway pizza \\
    subway & subway sushi \\
    Taco Bell & Taco Bell pizza \\
    Taco Bell & Taco Bell sushi \\
    Wendy's & Wendy's pizza \\
    Wendy's & Wendy's sushi \\ \hline
    \end{tabular}
    \caption{List of unsuitable objects.}
    \label{tab:removed target objects}
\end{table}

\subsection{Model Implementation}

We conduct the experiment on two models: Stable Diffusion v1.4~\cite{rombach2022high} (SD 1.4) and Stable Diffusion XL (SD XL). SD 1.4 has one text encoder, with dimension 768 and 16 cross-attention layers. SD XL has two text encoders, with dimensions 768 and 1280, 70 and 44 layers of cross-attention. 


\paragraph{Hyperparameter Sensitivity} 
\ee{} uses the same hyperparameters for both single and sequential edits across SD 1.4 and SD XL. \textbf{In contrast, TIME requires model-specific and edit-mode-specific tuning.} While TIME's recommended hyperparameter of 0.1 is effective for SD 1.4, it need to be adjusted significantly—from 0.1 to 10,000—to achieve reasonable performance with SD XL. \ee{}, however, demonstrate great robustness with consistent hyperparameters across models.

\subsection{Ablation Study}
To quantify the effect of the learning rate (lr) on \ee{}, We conducted an ablation study using 24 data samples, each samples generate 8 images. 
The results of this ablation study are presented in Table~\ref{tab:lr}. We select 0.001 for our experiment.

\begin{table}[ht]
\centering
\resizebox{0.9\columnwidth}{!}{
\begin{tabular}{l|lll}
\hline
lr & Efficacy & Generality & Specificity \\ \hline
0.1 & 78.65 & 77.08 & 53.96 \\
 & \scriptsize±5.37 & \scriptsize±6.04 & \scriptsize±5.09 \\
0.01 & 91.67 & 82.21 & 58.96 \\
 & \scriptsize±4.37 & \scriptsize±4.84 & \scriptsize±5.34 \\
0.001 & \underline{96.88} & \underline{84.87} & \underline{56.04} \\
 & \scriptsize±1.72 & \scriptsize±4.18 & \scriptsize±5.22 \\
0.0001 & 81.77 & 69.11 & 63.80 \\
 & \scriptsize±5.80 & \scriptsize ±5.44 & \scriptsize±4.82 \\
 \hline
\end{tabular}
}
\caption{Comparison of \ee{} with different values of learning rate. \% is omitted for clarity. Best results are marked with \underline{underline}.}
\label{tab:lr}
\end{table}

\subsection{Automatic Gender Method}
We design a new loss function to mitigate gender bias automatically. 
Details of the automatic method are shown in Equation~\ref{eq:inital}~\ref{eq:loss_mse}~\ref{eq:mse_csp}~\ref{eq:mse_sp}~\ref{eq:delta_bias rate MSE}~\ref{eq:alpha_value}. The auto method aims to modify the WTE of ``[profession]'' and seeks to mitigate gender bias in professions through a single edit.  

Results for the six professions after auto editing are shown in Table~\ref{tab:auto gender}, generating 10 images for each prompt. As anticipated, the auto method is able to adjust the gender bias but does not outperform manually adjusted settings. 

\begin{table}[ht]
\centering
\resizebox{0.9\columnwidth}{!}{
\begin{tabular}{cl|ccl}
\hline
 &  & Baseline & Manual & Auto \\ \hline
\multirow{6}{*}{\( F_p \)} & Hairdresser & 77.08 & 49.30 & 17.24 \\
 & CEO & 5.55 &  39.58 & 38.37 \\
 & Teacher & 80.55 &  57.63 & 53.33 \\
 & Lawyer & 29.86 &  55.84 & 66.67 \\
 & Housekeeper & 97.22 &  43.75 & 91.67 \\
 & Farmer & 3.47 &  55.56 & 23.33 \\ \hline
$\Delta_{p} (\downarrow)$ &  & 0.598 & 0.121 & 0.442 \\ \hline
\end{tabular}
}
\caption{Results of manual and auto edit on gender dataset. For $F_p$, ``{50}" represents the ideal debiased result (50 female, 50 male). $\Delta_{p}$ indicates the average deviation from 50, with smaller values reflecting a more neutral gender assumption. }
\label{tab:auto gender}
\end{table}

Define the ``[profession]'' as  \(\text{p}\), the ``[counter-stereotypical gender] [profession]'' as \( \text{csp}\), and the ``[stereotypical gender] [profession]'' as \( \text{sp} \). The corresponding last hidden states are represented as $\mathbf{h}{\text{p}}$, $\mathbf{h}{\text{csp}}$, and $\mathbf{h}_{\text{sp}}$. To mitigate gender bias, the model updates and optimizes the WTE vector associated with the source profession.

First, we initialize the embedding of target token $\mathrm{wte}_{\text{init}}$ as the average of three embeddings as shown in Eq.~\ref{eq:inital}
\begin{equation}
\mathrm{wte}_{\text{init}} = \frac{\mathrm{wte}_{p} + \mathrm{wte}_{\text{``female''}} + \mathrm{wte}_{\text{``male''}}}{3}
\label{eq:inital}
\end{equation}

Due to the varying biases among professions, we designed a reward-penalty loss to encourage the final embedding to move toward the direction of counter-stereotypical gender profession.

\begin{equation}
\text{\small Loss(p,sp,scp)} = \alpha^2 \cdot\text{MSE({\small p, csp})} + \left( \frac{1}{\alpha} \right)^2 \cdot\text{MSE({\small p, sp})}
\label{eq:loss_mse}
\end{equation}
where $\text{MSE(.)}$ stands for the MSE distance and is defined as follows:
\begin{equation}
\text{MSE({\small p,csp})} = \frac{1}{d} \sum_{i=1}^d \left( h_{\text{p}}, h_{\text{csp}}\right)^2
\label{eq:mse_csp}
\end{equation}
\begin{equation}
\text{MSE({\small p,sp})} = \frac{1}{d} \sum_{i=1}^d \left( h_{\text{p}}, h_{\text{sp}}\right)^2
\label{eq:mse_sp}
\end{equation}

We use $\alpha$ to control the contributions of either of the terms above for the final loss. The motivation is to balance the removal of gender bias and the adjustment of the target embedding. In particular, we determine the value of $\alpha$ considering the bias rate $\Delta$ of a specific profession using Eq.~\ref{eq:delta_bias rate MSE}:
\begin{equation}
\Delta(p,sp,csp) =  \frac{\|\text{MSE({p,sp})} - \text{MSE({p, csp})}\|}{0.5  \cdot \left( \text{MSE({p,sp})} + \text{MSE(p,csp)}\right)}
\label{eq:delta_bias rate MSE}
\end{equation}

We set $\alpha$ as in Eq.~\ref{eq:alpha_value}
\begin{equation}
\alpha = \max(\alpha_{\text{min}}, 10 \cdot \Delta) 
\label{eq:alpha_value}
\end{equation}
where the $\Delta$ is normalized, and \( \alpha_{\text{min}} \) represents the minimum weight to be set as 2.





\section{Additional Results}\label{sec:addition_results}
\begin{figure*}[ht]
    \centering
    \vspace{0.1cm}
    \textbf{\small Edit ``monster" to ``cookie monster"}
    \vspace{0.1cm}
    \\
    \begin{subfigure}[b]{0.32\textwidth}  
        \centering
        \includegraphics[width=0.3\linewidth]{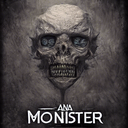}
        \hspace{-3pt} 
        \begin{overpic}[width=0.3\linewidth]{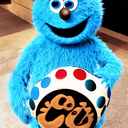}
            \put(-30, 25){\includegraphics[width=0.15\linewidth]{figs/right_arrow_1.png}}
        \end{overpic}
        \hspace{-3pt}
        \includegraphics[width=0.3\linewidth]{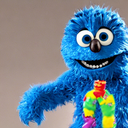}
        \caption{Efficacy: ``monster''}
    \end{subfigure}
    \hspace{-10pt}
    \begin{subfigure}[b]{0.32\textwidth}  
        \centering
        \includegraphics[width=0.3\linewidth]{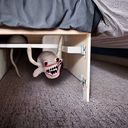}
        \hspace{-3pt}
        \begin{overpic}[width=0.3\linewidth]{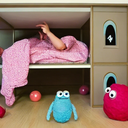}
            \put(-30, 25){\includegraphics[width=0.15\linewidth]{figs/right_arrow_1.png}}
        \end{overpic}
        \hspace{-3pt}
        \includegraphics[width=0.3\linewidth]{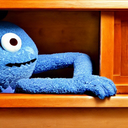}
        \caption{Generality: ``a monster under the bed''}
    \end{subfigure}
    \hspace{-10pt}
    \begin{subfigure}[b]{0.32\textwidth}  
        \centering
        \includegraphics[width=0.3\linewidth]{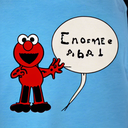}
        \hspace{-3pt}
        \begin{overpic}[width=0.3\linewidth]{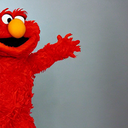}
            \put(-30, 25){\includegraphics[width=0.15\linewidth]{figs/right_arrow.png}}
        \end{overpic}
        \hspace{-3pt}
        \includegraphics[width=0.3\linewidth]{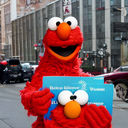}
        \caption{Specificity: ``elmo''}
    \end{subfigure}

    \vspace{0.1cm}
    \textbf{\small Edit ``dog" to ``poodle dog"}
    \vspace{0.1cm}
    \\
    \begin{subfigure}[b]{0.32\textwidth}  
        \centering
        \includegraphics[width=0.3\linewidth]{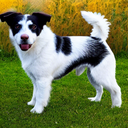}
        \hspace{-3pt} 
        \begin{overpic}[width=0.3\linewidth]{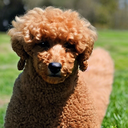}
            \put(-30, 25){\includegraphics[width=0.15\linewidth]{figs/right_arrow_1.png}}
        \end{overpic}
        \hspace{-3pt}
        \includegraphics[width=0.3\linewidth]{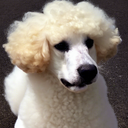}
        \caption{Efficacy: ``dog''}
    \end{subfigure}
    \hspace{-10pt}
    \begin{subfigure}[b]{0.32\textwidth}  
        \centering
        \includegraphics[width=0.3\linewidth]{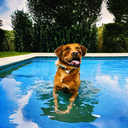}
        \hspace{-3pt}
        \begin{overpic}[width=0.3\linewidth]{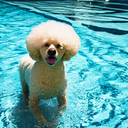}
            \put(-30, 25){\includegraphics[width=0.15\linewidth]{figs/right_arrow_1.png}}
        \end{overpic}
        \hspace{-3pt}
        \includegraphics[width=0.3\linewidth]{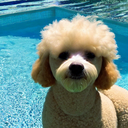}
        \caption{Generality: ``a photo of a dog in a pool''}
    \end{subfigure}
    \hspace{-10pt}
    \begin{subfigure}[b]{0.32\textwidth}  
        \centering
        \includegraphics[width=0.3\linewidth]{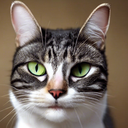}
        \hspace{-3pt}
        \begin{overpic}[width=0.3\linewidth]{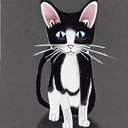}
            \put(-30, 25){\includegraphics[width=0.15\linewidth]{figs/right_arrow.png}}
        \end{overpic}
        \hspace{-3pt}
        \includegraphics[width=0.3\linewidth]{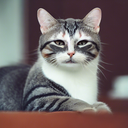}
        \caption{Specificity: ``a cat''}
    \end{subfigure}
    \caption{Illustration of Efficacy, Generality, and Specificity. Images are generated by \ee{}-edited Stable Diffusion v1.4.}
    \label{fig:sup_1_4_result}
\end{figure*}

\begin{figure*}[ht]
    \centering
    \vspace{0.1cm}
    \textbf{\small Edit ``ice cream" to ``pistachio ice cream"}
    \vspace{0.1cm}
    \\
    \begin{subfigure}[b]{0.32\textwidth}  
        \centering
        \includegraphics[width=0.3\linewidth]{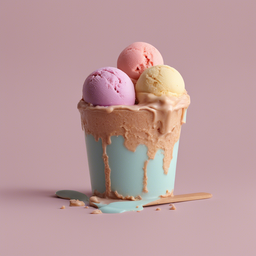}
        \hspace{-3pt} 
        \begin{overpic}[width=0.3\linewidth]{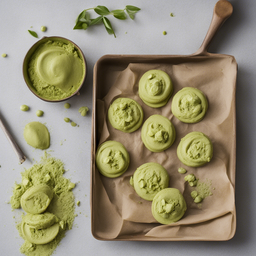}
            \put(-30, 25){\includegraphics[width=0.15\linewidth]{figs/right_arrow_1.png}}
        \end{overpic}
        \hspace{-3pt}
        \includegraphics[width=0.3\linewidth]{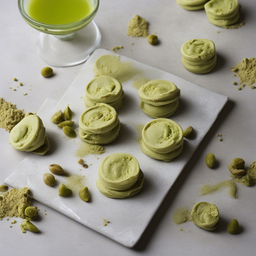}
        \caption{Efficacy: ``ice cream''}
    \end{subfigure}
    \hspace{-10pt}
    \begin{subfigure}[b]{0.32\textwidth}  
        \centering
        \includegraphics[width=0.3\linewidth]{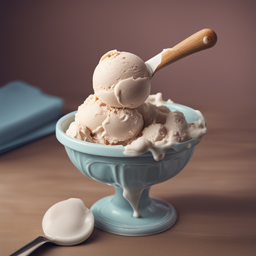}
        \hspace{-3pt}
        \begin{overpic}[width=0.3\linewidth]{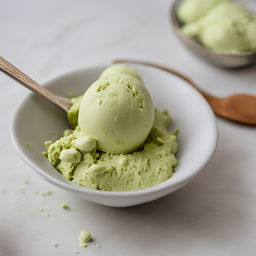}
            \put(-30, 25){\includegraphics[width=0.15\linewidth]{figs/right_arrow_1.png}}
        \end{overpic}
        \hspace{-3pt}
        \includegraphics[width=0.3\linewidth]{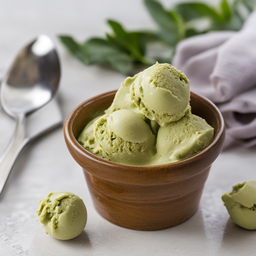}
        \caption{Generality: ``a scoop of ice cream''}
    \end{subfigure}
    \hspace{-10pt}
    \begin{subfigure}[b]{0.32\textwidth}  
        \centering
        \includegraphics[width=0.3\linewidth]{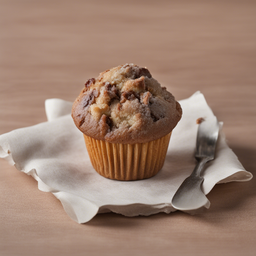}
        \hspace{-3pt}
        \begin{overpic}[width=0.3\linewidth]{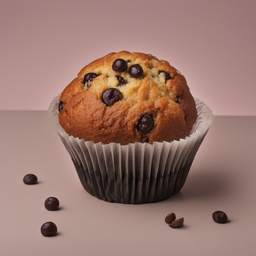}
            \put(-30, 25){\includegraphics[width=0.15\linewidth]{figs/right_arrow.png}}
        \end{overpic}
        \hspace{-3pt}
        \includegraphics[width=0.3\linewidth]{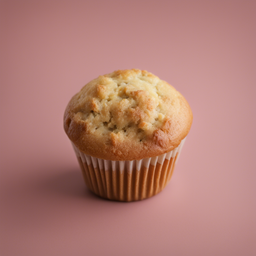}
        \caption{Specificity: ``a photo of a muffin''}
    \end{subfigure}

    \vspace{0.1cm}
    \textbf{\small Edit ``dog" to ``Chihuahua dog"}
    \vspace{0.1cm}
    \\
    \begin{subfigure}[b]{0.32\textwidth}  
        \centering
        \includegraphics[width=0.3\linewidth]{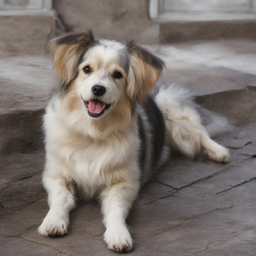}
        \hspace{-3pt} 
        \begin{overpic}[width=0.3\linewidth]{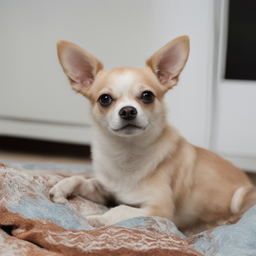}
            \put(-30, 25){\includegraphics[width=0.15\linewidth]{figs/right_arrow_1.png}}
        \end{overpic}
        \hspace{-3pt}
        \includegraphics[width=0.3\linewidth]{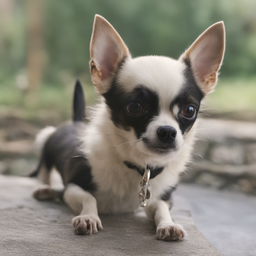}
        \caption{Efficacy: ``dog''}
    \end{subfigure}
    \hspace{-10pt}
    \begin{subfigure}[b]{0.32\textwidth}  
        \centering
        \includegraphics[width=0.3\linewidth]{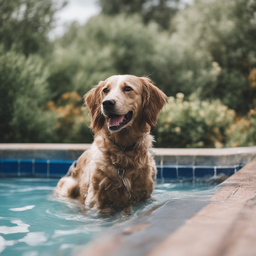}
        \hspace{-3pt}
        \begin{overpic}[width=0.3\linewidth]{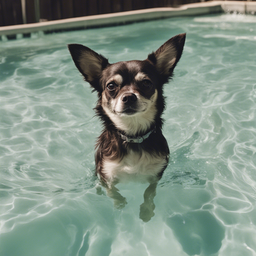}
            \put(-30, 25){\includegraphics[width=0.15\linewidth]{figs/right_arrow_1.png}}
        \end{overpic}
        \hspace{-3pt}
        \includegraphics[width=0.3\linewidth]{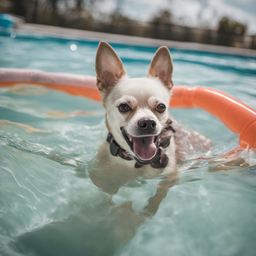}
        \caption{Generality: ``a photo of a dog in a pool''}
    \end{subfigure}
    \hspace{-10pt}
    \begin{subfigure}[b]{0.32\textwidth}  
        \centering
        \includegraphics[width=0.3\linewidth]{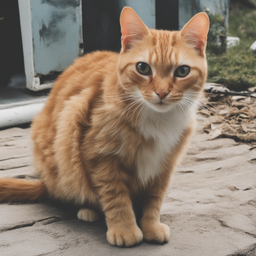}
        \hspace{-3pt}
        \begin{overpic}[width=0.3\linewidth]{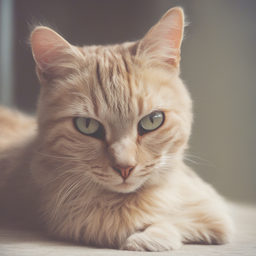}
            \put(-30, 25){\includegraphics[width=0.15\linewidth]{figs/right_arrow.png}}
        \end{overpic}
        \hspace{-3pt}
        \includegraphics[width=0.3\linewidth]{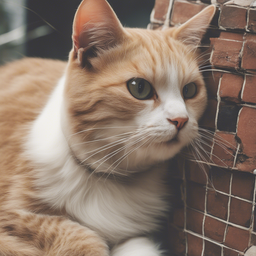}
        \caption{Specificity: ``a cat''}
    \end{subfigure}
    \caption{Illustration of Efficacy, Generality, and Specificity. Images are generated by \ee{}-edited Stable Diffusion XL.}
    \label{fig:sup_XL_result}
\end{figure*}

\begin{figure*}[b]
    \centering
    \textbf{\large Edit ``plum" to ``yellow plum"}
    \vspace{0.1cm}
    \\    

    \textbf{\small  \ee{} Single Edit on Stable Diffusion v1.4}
    \vspace{0.1cm}
    \\
    \begin{subfigure}[b]{0.32\textwidth}  
        \centering
        \includegraphics[width=0.3\linewidth]{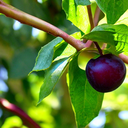}
        \hspace{-3pt} 
        \begin{overpic}[width=0.3\linewidth]{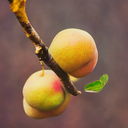}
            \put(-30, 25){\includegraphics[width=0.15\linewidth]{figs/right_arrow_1.png}}
        \end{overpic}
        \hspace{-3pt}
        \includegraphics[width=0.3\linewidth]{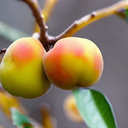}
        \caption{Efficacy: ``plum''}
    \end{subfigure}
    \hspace{-10pt}
    \begin{subfigure}[b]{0.32\textwidth}  
        \centering
        \includegraphics[width=0.3\linewidth]{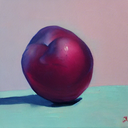}
        \hspace{-3pt}
        \begin{overpic}[width=0.3\linewidth]{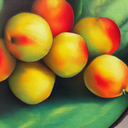}
            \put(-30, 25){\includegraphics[width=0.15\linewidth]{figs/right_arrow_1.png}}
        \end{overpic}
        \hspace{-3pt}
        \includegraphics[width=0.3\linewidth]{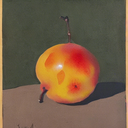}
        \caption{Generality: ``a painting of plum''}
    \end{subfigure}
    \hspace{-10pt}
    \begin{subfigure}[b]{0.32\textwidth}  
        \centering
        \includegraphics[width=0.3\linewidth]{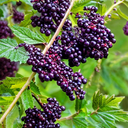}
        \hspace{-3pt}
        \begin{overpic}[width=0.3\linewidth]{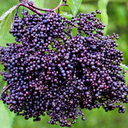}
            \put(-30, 25){\includegraphics[width=0.15\linewidth]{figs/right_arrow.png}}
        \end{overpic}
        \hspace{-3pt}
        \includegraphics[width=0.3\linewidth]{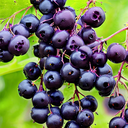}
        \caption{Specificity: ``elderberries''}
    \end{subfigure}

    \textbf{\small  TIME Single Edit on Stable Diffusion v1.4}
    \vspace{0.1cm}
    \\
    \begin{subfigure}[b]{0.32\textwidth}  
        \centering
        \includegraphics[width=0.3\linewidth]{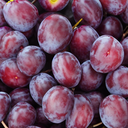}
        \hspace{-3pt} 
        \begin{overpic}[width=0.3\linewidth]{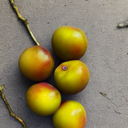}
            \put(-30, 25){\includegraphics[width=0.15\linewidth]{figs/right_arrow_1.png}}
        \end{overpic}
        \hspace{-3pt}
        \includegraphics[width=0.3\linewidth]{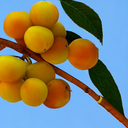}
        \caption{Efficacy: ``plum''}
    \end{subfigure}
    \hspace{-10pt}
    \begin{subfigure}[b]{0.32\textwidth}  
        \centering
        \includegraphics[width=0.3\linewidth]{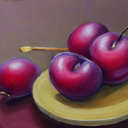}
        \hspace{-3pt}
        \begin{overpic}[width=0.3\linewidth]{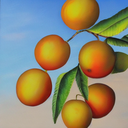}
            \put(-30, 25){\includegraphics[width=0.15\linewidth]{figs/right_arrow_1.png}}
        \end{overpic}
        \hspace{-3pt}
        \includegraphics[width=0.3\linewidth]{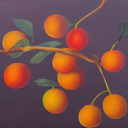}
        \caption{Generality: ``a painting of plum''}
    \end{subfigure}
    \hspace{-10pt}
    \begin{subfigure}[b]{0.32\textwidth}  
        \centering
        \includegraphics[width=0.3\linewidth]{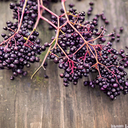}
        \hspace{-3pt}
        \begin{overpic}[width=0.3\linewidth]{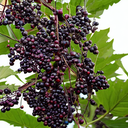}
            \put(-30, 25){\includegraphics[width=0.15\linewidth]{figs/right_arrow.png}}
        \end{overpic}
        \hspace{-3pt}
        \includegraphics[width=0.3\linewidth]{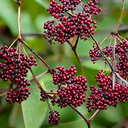}
        \caption{Specificity: ``elderberries''}
    \end{subfigure}

    \textbf{\small  \ee{} Sequential Edit on Stable Diffusion v1.4}
    \vspace{0.1cm}
    \\
    \begin{subfigure}[b]{0.32\textwidth}  
        \centering
        \includegraphics[width=0.3\linewidth]{figs/Supplementary/yellow_plum/11.png}
        \hspace{-3pt} 
        \begin{overpic}[width=0.3\linewidth]{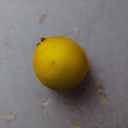}
            \put(-30, 25){\includegraphics[width=0.15\linewidth]{figs/right_arrow_1.png}}
        \end{overpic}
        \hspace{-3pt}
        \includegraphics[width=0.3\linewidth]{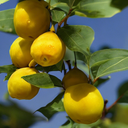}
        \caption{Efficacy: ``plum''}
    \end{subfigure}
    \hspace{-10pt}
    \begin{subfigure}[b]{0.32\textwidth}  
        \centering
        \includegraphics[width=0.3\linewidth]{figs/Supplementary/yellow_plum/21.png}
        \hspace{-3pt}
        \begin{overpic}[width=0.3\linewidth]{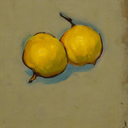}
            \put(-30, 25){\includegraphics[width=0.15\linewidth]{figs/right_arrow_1.png}}
        \end{overpic}
        \hspace{-3pt}
        \includegraphics[width=0.3\linewidth]{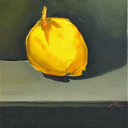}
        \caption{Generality: ``a painting of plum''}
    \end{subfigure}
    \hspace{-10pt}
    \begin{subfigure}[b]{0.32\textwidth}  
        \centering
        \includegraphics[width=0.3\linewidth]{figs/Supplementary/yellow_plum/31.png}
        \hspace{-3pt}
        \begin{overpic}[width=0.3\linewidth]{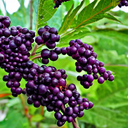}
            \put(-30, 25){\includegraphics[width=0.15\linewidth]{figs/right_arrow.png}}
        \end{overpic}
        \hspace{-3pt}
        \includegraphics[width=0.3\linewidth]{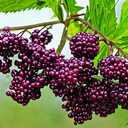}
        \caption{Specificity: ``elderberries''}
    \end{subfigure}

    \textbf{\small  TIME Sequential Edit on Stable Diffusion v1.4}
    \vspace{0.1cm}
    \\
    \begin{subfigure}[b]{0.32\textwidth}  
        \centering
        \includegraphics[width=0.3\linewidth]{figs/Supplementary/plum/1.4-TIME/base/baseline.png}
        \hspace{-3pt} 
        \begin{overpic}[width=0.3\linewidth]{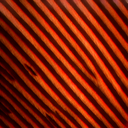}
            \put(-30, 25){\includegraphics[width=0.15\linewidth]{figs/right_arrow_1.png}}
        \end{overpic}
        \hspace{-3pt}
        \includegraphics[width=0.3\linewidth]{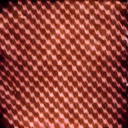}
        \caption{Efficacy: ``plum''}
    \end{subfigure}
    \hspace{-10pt}
    \begin{subfigure}[b]{0.32\textwidth}  
        \centering
        \includegraphics[width=0.3\linewidth]{figs/Supplementary/plum/1.4-TIME/a_painting_of_plum/baseline.png}
        \hspace{-3pt}
        \begin{overpic}[width=0.3\linewidth]{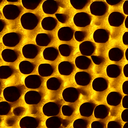}
            \put(-30, 25){\includegraphics[width=0.15\linewidth]{figs/right_arrow_1.png}}
        \end{overpic}
        \hspace{-3pt}
        \includegraphics[width=0.3\linewidth]{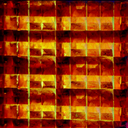}
        \caption{Generality: ``a painting of plum''}
    \end{subfigure}
    \hspace{-10pt}
    \begin{subfigure}[b]{0.32\textwidth}  
        \centering
        \includegraphics[width=0.3\linewidth]{figs/Supplementary/plum/1.4-TIME/elderberry/baseline.png}
        \hspace{-3pt}
        \begin{overpic}[width=0.3\linewidth]{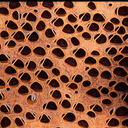}
            \put(-30, 25){\includegraphics[width=0.15\linewidth]{figs/right_arrow.png}}
        \end{overpic}
        \hspace{-3pt}
        \includegraphics[width=0.3\linewidth]{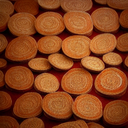}
        \caption{Specificity: ``elderberries''}
    \end{subfigure}
    \\
    

    \centering
    
    \textbf{\small \ee{} Single Edit on Stable Diffusion XL}
    \vspace{0.1cm}
    \\
    \begin{subfigure}[b]{0.32\textwidth}  
        \centering
        \includegraphics[width=0.3\linewidth]{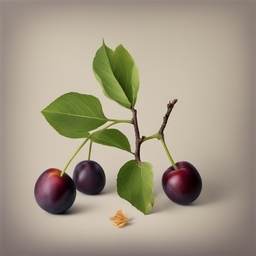}
        \hspace{-3pt} 
        \begin{overpic}[width=0.3\linewidth]{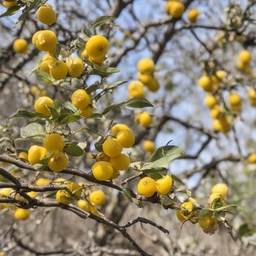}
            \put(-30, 25){\includegraphics[width=0.15\linewidth]{figs/right_arrow_1.png}}
        \end{overpic}
        \hspace{-3pt}
        \includegraphics[width=0.3\linewidth]{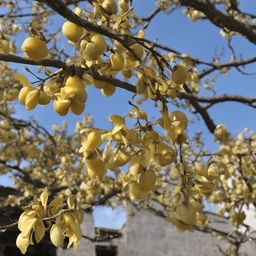}
        \caption{Efficacy: ``plum''}
    \end{subfigure}
    \hspace{-10pt}
    \begin{subfigure}[b]{0.32\textwidth}  
        \centering
        \includegraphics[width=0.3\linewidth]{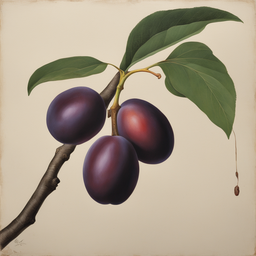}
        \hspace{-3pt}
        \begin{overpic}[width=0.3\linewidth]{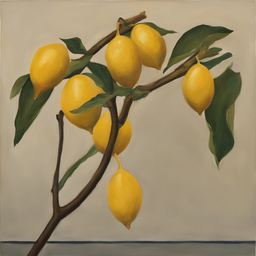}
            \put(-30, 25){\includegraphics[width=0.15\linewidth]{figs/right_arrow_1.png}}
        \end{overpic}
        \hspace{-3pt}
        \includegraphics[width=0.3\linewidth]{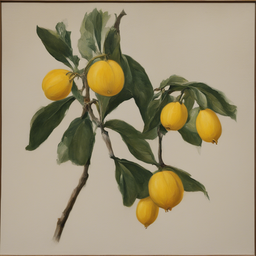}
        \caption{Generality: ``a painting of plum''}
    \end{subfigure}
    \hspace{-10pt}
    \begin{subfigure}[b]{0.32\textwidth}  
        \centering
        \includegraphics[width=0.3\linewidth]{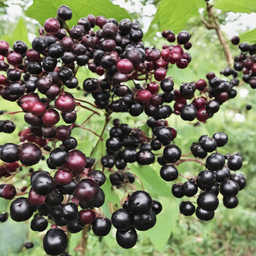}
        \hspace{-3pt}
        \begin{overpic}[width=0.3\linewidth]{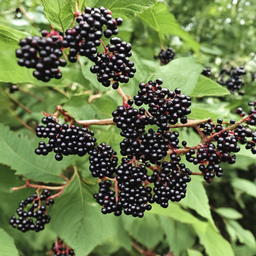}
            \put(-30, 25){\includegraphics[width=0.15\linewidth]{figs/right_arrow.png}}
        \end{overpic}
        \hspace{-3pt}
        \includegraphics[width=0.3\linewidth]{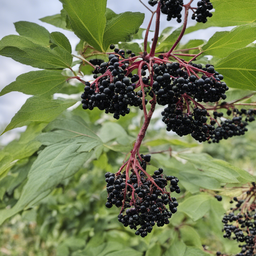}
        \caption{Specificity: ``elderberries''}
    \end{subfigure}
    
    \centering
    
    \textbf{\small TIME Single Edit on Stable Diffusion XL}
    \vspace{0.1cm}
    \\
    \begin{subfigure}[b]{0.32\textwidth}  
        \centering
        \includegraphics[width=0.3\linewidth]{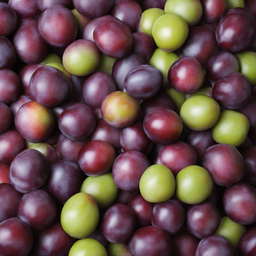}
        \hspace{-3pt} 
        \begin{overpic}[width=0.3\linewidth]{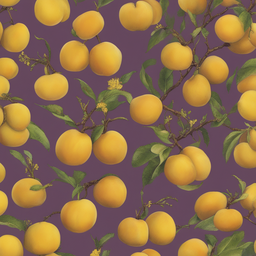}
            \put(-30, 25){\includegraphics[width=0.15\linewidth]{figs/right_arrow_1.png}}
        \end{overpic}
        \hspace{-3pt}
        \includegraphics[width=0.3\linewidth]{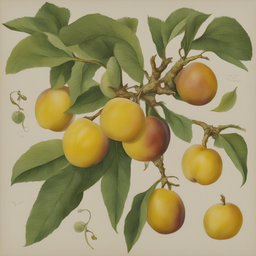}
        \caption{Efficacy: ``plum''}
    \end{subfigure}
    \hspace{-10pt}
    \begin{subfigure}[b]{0.32\textwidth}  
        \centering
        \includegraphics[width=0.3\linewidth]{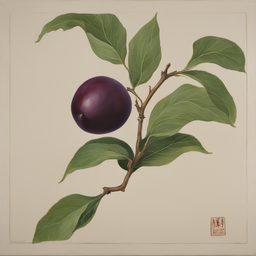}
        \hspace{-3pt}
        \begin{overpic}[width=0.3\linewidth]{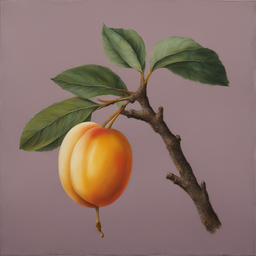}
            \put(-30, 25){\includegraphics[width=0.15\linewidth]{figs/right_arrow_1.png}}
        \end{overpic}
        \hspace{-3pt}
        \includegraphics[width=0.3\linewidth]{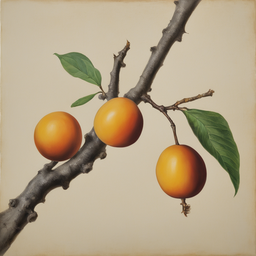}
        \caption{Generality: ``a painting of plum''}
    \end{subfigure}
    \hspace{-10pt}
    \begin{subfigure}[b]{0.32\textwidth}  
        \centering
        \includegraphics[width=0.3\linewidth]{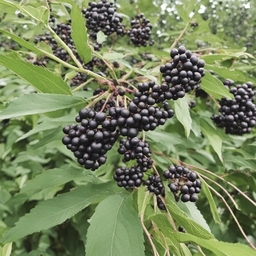}
        \hspace{-3pt}
        \begin{overpic}[width=0.3\linewidth]{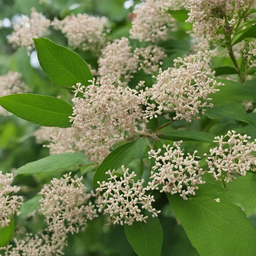}
            \put(-30, 25){\includegraphics[width=0.15\linewidth]{figs/right_arrow.png}}
        \end{overpic}
        \hspace{-3pt}
        \includegraphics[width=0.3\linewidth]{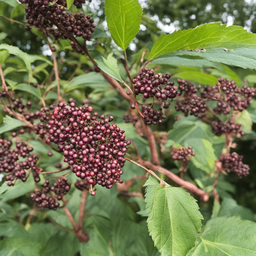}
        \caption{Specificity: ``elderberries''}
    \end{subfigure}
    
    \centering
    
    \textbf{\small \ee{} Sequential Edit on Stable Diffusion XL}
    \vspace{0.1cm}
    \\
    \begin{subfigure}[b]{0.32\textwidth}  
        \centering
        \includegraphics[width=0.3\linewidth]{figs/Supplementary/plum/xl_single.jpg}
        \hspace{-3pt} 
        \begin{overpic}[width=0.3\linewidth]{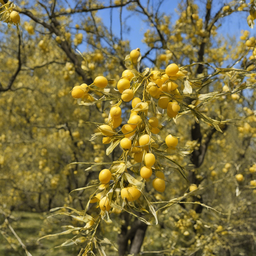}
            \put(-30, 25){\includegraphics[width=0.15\linewidth]{figs/right_arrow_1.png}}
        \end{overpic}
        \hspace{-3pt}
        \includegraphics[width=0.3\linewidth]{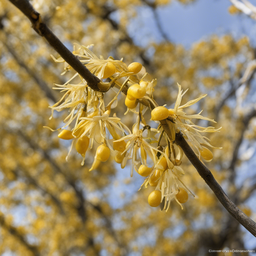}
        \caption{Efficacy: ``plum''}
    \end{subfigure}
    \hspace{-10pt}
    \begin{subfigure}[b]{0.32\textwidth}  
        \centering
        \includegraphics[width=0.3\linewidth]{figs/Supplementary/pics/plum_yellow_plum/a_painting_of_plum/baseline.png}
        \hspace{-3pt}
        \begin{overpic}[width=0.3\linewidth]{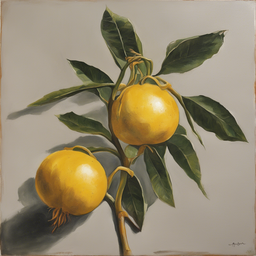}
            \put(-30, 25){\includegraphics[width=0.15\linewidth]{figs/right_arrow_1.png}}
        \end{overpic}
        \hspace{-3pt}
        \includegraphics[width=0.3\linewidth]{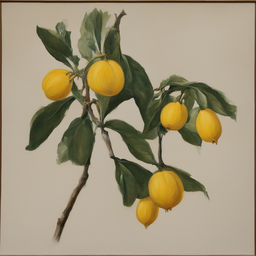}
        \caption{Generality: ``a painting of plum''}
    \end{subfigure}
    \hspace{-10pt}
    \begin{subfigure}[b]{0.32\textwidth}  
        \centering
        \includegraphics[width=0.3\linewidth]{figs/Supplementary/pics/plum_yellow_plum/elderberries/baseline.png}
        \hspace{-3pt}
        \begin{overpic}[width=0.3\linewidth]{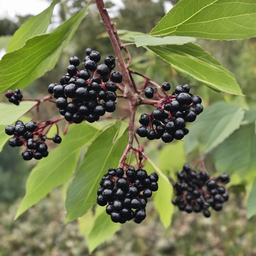}
            \put(-30, 25){\includegraphics[width=0.15\linewidth]{figs/right_arrow.png}}
        \end{overpic}
        \hspace{-3pt}
        \includegraphics[width=0.3\linewidth]{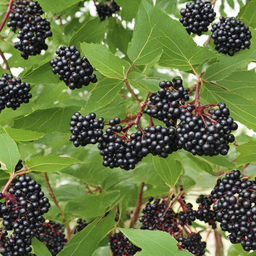}
        \caption{Specificity: ``elderberries''}
    \end{subfigure}

    \centering
    
    \textbf{\small TIME Sequential Edit on Stable Diffusion XL}
    \vspace{0.1cm}
    \\
    \begin{subfigure}[b]{0.32\textwidth}  
        \centering
        \includegraphics[width=0.3\linewidth]{figs/Supplementary/plum/xl-TIME/base/baseline.png}
        \hspace{-3pt} 
        \begin{overpic}[width=0.3\linewidth]{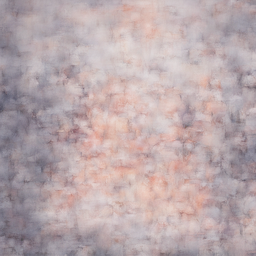}
            \put(-30, 25){\includegraphics[width=0.15\linewidth]{figs/right_arrow_1.png}}
        \end{overpic}
        \hspace{-3pt}
        \includegraphics[width=0.3\linewidth]{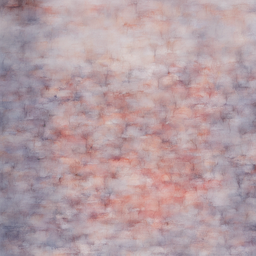}
        \caption{Efficacy: ``plum''}
    \end{subfigure}
    \hspace{-10pt}
    \begin{subfigure}[b]{0.32\textwidth}  
        \centering
        \includegraphics[width=0.3\linewidth]{figs/Supplementary/plum/xl-TIME/a_painting_of_plum/baseline.png}
        \hspace{-3pt}
        \begin{overpic}[width=0.3\linewidth]{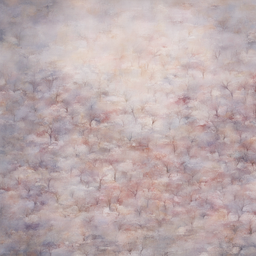}
            \put(-30, 25){\includegraphics[width=0.15\linewidth]{figs/right_arrow_1.png}}
        \end{overpic}
        \hspace{-3pt}
        \includegraphics[width=0.3\linewidth]{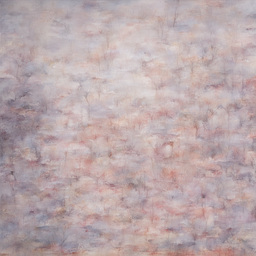}
        \caption{Generality: ``a painting of plum''}
    \end{subfigure}
    \hspace{-10pt}
    \begin{subfigure}[b]{0.32\textwidth}  
        \centering
        \includegraphics[width=0.3\linewidth]{figs/Supplementary/plum/xl_elderberry.jpg}
        \hspace{-3pt}
        \begin{overpic}[width=0.3\linewidth]{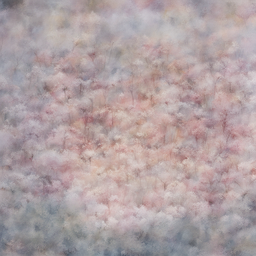}
            \put(-30, 25){\includegraphics[width=0.15\linewidth]{figs/right_arrow.png}}
        \end{overpic}
        \hspace{-3pt}
        \includegraphics[width=0.3\linewidth]{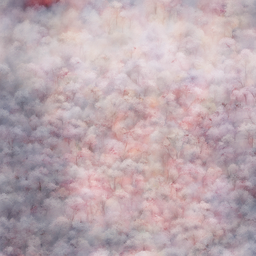}
        \caption{Specificity: ``elderberries''}
    \end{subfigure}
    \caption{Comparison of \ee{} and TIME on SD v1.4 and SD XL for single and sequential edits.}
    \label{fig:sup_comparsion_plum}
\end{figure*}

We present additional qualitative results of \ee{}. Figure ~\ref{fig:sup_1_4_result} illustrates the generalization and specificity of \ee{} on SD v1.4. Figure ~\ref{fig:sup_XL_result} on SD XL.
Figure ~\ref{fig:sup_comparsion_plum} is a comparison of \ee{} and TIME performance on SD v1.4 and SD XL.

We also compare our method with ReFACT~\cite{arad-etal-2024-refact}, an approach for editing factual associations. While ReFACT modifies the multi-layer perceptron (MLP) layers, our method focuses solely on editing the WTE parameters. 

For evaluation, we use \ee{} and ReFACT to generate eight images per prompt from the RoAD dataset. We use the CLIP ViT-B/32~\cite{radford2021learning} model as a zero-shot text-based classifier. 

As shown in Table~\ref{tab:road_comparsion}, \ee{} significantly outperforms ReFACT in terms of specificity, which is expected since \ee{} does not modify the model internally. For efficacy and generality, both ReFACT and \ee{} achieve comparable performance within the margin of standard error. 
Additionally, we compare computational efficiency using the average edit time. \ee{} is significantly faster, requiring only an average of 0.37 seconds to edit an original object into a new one.

\begin{table}[ht]
\resizebox{\columnwidth}{!}{
\begin{tabular}{lcccc}
\hline
\multicolumn{5}{c}{RoAD dataset on SD 1.4} \\ \hline
\multicolumn{1}{l|}{Method} & Oracle & Baseline & ReFACT & EmbEdit \\ \hline
\multicolumn{1}{l|}{Efficacy} & 99.72 & 1.99 & 92.26 & 91.19 \\
\multicolumn{1}{l|}{} & \scriptsize±0.28 & \scriptsize±1.46 & \scriptsize±2.37 & \scriptsize±3.19 \\
\multicolumn{1}{l|}{Generality} & 96.76 & 7.33 & 83.51 & 82.44 \\
\multicolumn{1}{l|}{} & \scriptsize±0.82 & \scriptsize±1.95 & \scriptsize±3.32 & \scriptsize±3.41 \\
\multicolumn{1}{l|}{Specificity} & 97.54 & 97.54 & 80.40 & \textbf{88.18} \\
\multicolumn{1}{l|}{} & \scriptsize±0.79 & \scriptsize±0.79 & \scriptsize±2.97 & \scriptsize±2.15 \\ \hline
\multicolumn{1}{l|}{FID} & 40.13 & 40.13 & 41.93 & 40.66 \\
\multicolumn{1}{l|}{CLIP Score} & 31.17 & 31.17 & 30.59 & 31.04 \\
\multicolumn{1}{l|}{Average edit time} & - & - & 89.75s & \textbf{0.37}s
\\ \hline
\end{tabular}%
}
\caption{\ee{} and ReFACT sequential edit performance and generative quality comparison on SD v1.4. 
Sequential editing means applying all edits one after another on a single set of weights, without reloading them.
Best results of \ee{} is highlighted in bold.
}
\label{tab:road_comparsion}
\end{table}
Our method does not support the retention the multi-hop reasoning on the target object \cite{yang-etal-2024-large-language-models}. We selectively experiments with 10 of this examples, one of the result is shown on Figure~\ref{fig:tomhanks}:

\begin{figure}[h!]
\centering
    \vspace{0.1cm}
    \textbf{\small Edit ``Jason Alexander'' to ``Tom Hanks''}
    \vspace{0.1cm}
    \\
    \begin{subfigure}[b]{0.45\textwidth}
    \begin{subfigure}{0.30\linewidth}
        \centering
        \includegraphics[width=0.9\linewidth]{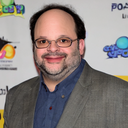}
    \end{subfigure}
    \hfill
    \begin{subfigure}{0.30\linewidth}
        \centering
        \includegraphics[width=0.9\linewidth]{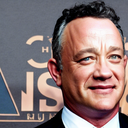}
    \end{subfigure}
    \hfill
    \begin{subfigure}{0.30\linewidth}
        \centering
        \includegraphics[width=0.9\linewidth]{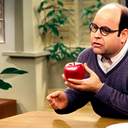}
    \end{subfigure}
    \end{subfigure}
    \caption{An example of editing ``Jason Alexander'' to ``Tom Hanks'': P1 is the unedited baseline with prompt ``Jason Alexander'', P2 is a successful edit with prompt ``Jason Alexander'' and shown the actor Tom Hanks, while P3 is a multi-hop test with ``George Costanza eating an apple''.}
    \label{fig:tomhanks}
\end{figure}


\section{Ethical Considerations \& Safety}

\ee{} allows model editing with extremely low computational resources, which could be misused to spread misinformation or offensive content. However, given extensive research on mitigating harmful representations~\cite{bolukbasi2016man, bianchi2023easily}, we believe the benefits of sharing our method outweigh the risks.

Additionally, we place a high emphasis on the transparency of our research process to ensure that other researchers can understand and replicate our experiments. All tool versions, experimental setups, and parameter configurations are detailed in the appendix and the relevant resources and data are provided through a publicly accessible code repository. This not only facilitates scientific communication and collaboration but also aids in the verification of results and further research.

\onecolumn

\twocolumn